\documentclass[runningheads]{llncs}

\usepackage{latexsym}
\usepackage{amssymb}
\usepackage{amsmath}
\usepackage{booktabs}
\usepackage{enumitem}
\usepackage{graphicx}
\usepackage{color}

\usepackage{times}
\usepackage{soul}
\usepackage{url}
\usepackage[hidelinks]{hyperref}
\usepackage[utf8]{inputenc}
\usepackage{graphicx}
\usepackage{amsmath}
\usepackage{booktabs}
\usepackage{algorithm}
\usepackage{algpseudocode}
\usepackage{amsmath}
\usepackage[switch]{lineno}
\usepackage{bm}
\usepackage{multirow}
\providecommand{\orcidID}[1]{}

\usepackage[utf8]{inputenc}
\usepackage{graphicx}
\usepackage{float}
\usepackage{algorithm}
\usepackage{algpseudocode}
\usepackage{tikz}
\usetikzlibrary{decorations.markings}
\usetikzlibrary{arrows}
\usetikzlibrary{shapes}
\usetikzlibrary{positioning}
\usepackage{pgfplots}
\pgfplotsset{compat=1.10}
\usetikzlibrary{shapes.geometric,arrows,fit,matrix,positioning}
\tikzset
{
    treenode/.style = {circle, draw=black, align=center, minimum size=1.1cm},
}

\renewcommand{\vec}[1]{\bm{#1}}
\def\cnf{{\tt CNF}}

\def\dDNNF{{\tt d-DNNF}}

\newcommand{\BibTeX}{B\kern-.05em{\sc i\kern-.025em b}\kern-.08em\TeX}

\begin{document}


\title{A Rectification-Based Approach for\\ Distilling Boosted Trees into Decision Trees}

\author{Gilles Audemard\inst{1}\orcidID{0000-0003-2604-9657} \and
Sylvie Coste-Marquis\inst{1}\orcidID{0000-0003-4742-4858} \and Pierre Marquis\inst{1,2}\orcidID{0000-0002-7979-6608} \and 
Mehdi Sabiri\inst{1}\orcidID{0009-0003-8642-9755} \and Nicolas Szczepanski\inst{1}\orcidID{0000-0001-7553-5657}}
\authorrunning{G. Audemard et al.}
%
\institute{Univ. Artois, CNRS, CRIL\\
\email{name@cril.fr}\\
\url{http://www.cril.fr} \and
Institut Universitaire de France 
}


\maketitle 

\begin{abstract}
We present a new approach for distilling boosted trees into decision trees, in the objective of generating an ML model offering an acceptable compromise in terms of predictive performance and interpretability. We explain how the correction approach called rectification 
can be used to implement such a distillation process. We show empirically that this approach provides interesting results, in comparison with an approach to distillation achieved by retraining the model.
\end{abstract}

\section{Introduction}

Applications of machine learning (ML) have flourished over the last decade, marked by the emergence of ML-based AI systems offering increasingly higher levels of predictive performance.
Nevertheless, there is a wide range of critical applications of such systems (for example, in the health domain or in the legal domain) in which more than predictions are expected: users must be allowed to interpret the results obtained, receive explanations for the predictions that have been made by the system and, when possible, correct the prediction errors. 

The field of “eXplainable AI (XAI)” was born a few years ago \cite{DBLP:conf/iui/Gunning19} with the goal to get AI systems that are more interpretable.
More precisely, DARPA ({\it Defense Advanced Research Projects Agency}), at the origin of the term "XAI", put forward the following objectives: \emph{''to provide users with explanations that allow them to understand the forces and the overall weaknesses of the system in question, which allow them to understand how it will behave in the future, or even to correct the system's errors''}.

Unfortunately, the most accurate ML models are difficult to interpret, and vice versa, the most interpretable models are not always very accurate  \cite{DBLP:journals/inffus/ArrietaRSBTBGGM20}. A precision-interpretability compromise should therefore be considered  when an ML-based AI system is to be used in a critical application.
What makes the problem hard enough is that interpretability is a domain-specific notion \cite{DBLP:journals/natmi/Rudin19}. Especially, the interpretability of a ML model can be evaluated from several points of view. These include the clarity of the method used to construct the model, the number of parameters, the structure or size of the model, the possibility of extracting from the model simple classification rules or explanations, etc. (see e.g., \cite{DBLP:journals/corr/abs-1907-06831,Zhouetal21,DBLP:journals/inffus/ViloneL21,DBLP:conf/ijcai/AmgoudB22,Nautaetal23}).

\emph{Decision trees} constitute an ML model that is not always very accurate in practice because of its algorithmic unstability, but 
is often considered as \emph{interpretable by design} \cite{Molnar19}.
Indeed, from a decision tree, it is possible to derive in linear time an equivalent set of classification rules (corresponding to the paths in the tree).
This makes the tree globally interpretable, provided that the rules are not too numerous and too large (i.e., the tree is not too deep).
More generally, past work showed that decision trees are \emph{computationally intelligible} in the sense that they support in polynomial time a wide range of explanation queries and verification queries, the answers to which can be used by the user to \emph{decide} whether to trust the predictions made \cite{DBLP:conf/kr/AudemardBBKLM21,DBLP:journals/dke/AudemardBBKLM22}.
In practice, the answers to those queries can be derived in a reasonable amount of time, even when the tree 
contains a very large number of nodes and/or have branches that are too deep for being considered as globally interpretable or human comprehensible. 
Conversely, deep neural networks and boosted trees are other ML models that often exhibit an impressive predictive performance but can hardly be viewed as interpretable, even from the point of view of their computational intelligibility \cite{DBLP:journals/dke/AudemardBBKLM22}. 

\medskip

Model \emph{distillation} \cite{hinton2015distilling,DBLP:journals/corr/abs-1711-09784,DBLP:journals/ijcv/GouYMT21} is a method for building a “simpler” target ML model than the  source ML model considered as input. The source model and the target model can be from the same family, but not necessarily. The desired simplicity can be expressed in terms of number of parameters, 
and various objectives can be considered  such as reducing the amount of memory required to run the model, reducing the time needed to get predictions, providing explanations \cite{asadulaev2019interpretable,DBLP:conf/bionlp/Wood-DoughtyCD22}, etc.
Distillation constitutes, in particular, a possible approach \emph{to achieve a good accuracy/inte\-rpretability compromise}, by making it possible to generate a sufficiently precise, yet interpretable ML model $I$ from a poorly interpretable but accurate ML model $P$.


\medskip
In this paper, we focus on an \emph{incremental distillation process}, where the aim is to \emph{correct} an initial ML model $I$ for binary classification, that is quite interpretable but not very accurate, using another ML model $P$ for binary classification, that is quite precise but not interpretable. 
Here $I$ is a \emph{decision tree} \cite{DBLP:books/wa/BreimanFOS84,DBLP:journals/ml/Quinlan86},
and $P$ is a \emph{boosted tree} \cite{DBLP:journals/jcss/FreundS97}.
Our objective is to correct $I$ in an incremental way to make it logically closer to $P$ at each correction step (i.e., to increase the number of instances $\vec x$ such that $I(\vec x) = P(\vec x)$), while preserving the computational intelligibility offered by the decision tree model. Note that improving the predictive performance of $P$ by combining $I$ with $P$ would be a different story. 
Especially, since $P$ is used as an oracle in our approach, when its predictive performance is bad, so will be the predictive performance of $I$ at the end of the correction process.

In our work, the benefits that are expected from the distillation of  $P$ into $I$ are from the XAI side. In practice, many efficient XAI algorithms can be leveraged when dealing with decision trees \cite{DBLP:conf/kr/AudemardBBKLM21}, while only a few XAI algorithms have been implemented and are available online for boosted trees. Furthermore, because of the computational complexity of XAI queries for boosted trees, such XAI algorithms do not scale up well. Thus, when the goal is to compute a subset-minimal abductive explanation (aka a sufficient reason) \cite{IgnatievNM19,DarwicheH20} for an instance given a boosted tree $P$, there is no guarantee that any state-of-the-art algorithm (like the one presented in \cite{DBLP:conf/aaai/IgnatievIS022}) will be able to return such an explanation in a reasonable amount of time, especially when $P$ contains many trees over a large number of features. Furthermore, the success of the computation in due time
of a sufficient reason for an instance $\vec x$ given $P$  may depend heavily on the instance $\vec x$ one starts with. In contrast, the critical part in the computation of a sufficient reason  for $\vec x$ given $P$ through the distillation of $P$ into $I$ is the distillation process itself: once $I$ has been computed from $P$, even when $I$ is large enough,  computing a sufficient reason for $\vec x$ given $I$  (and answering other XAI queries) turns out to be feasible in a reasonable amount of time whatever the instance at hand. This kind of behaviour is quite standard when \emph{compiling representations} \cite{CadoliDonini97}: $P$, which can be viewed as the (so-called) fixed part of the compilation problem  is compiled (i.e., distilled) into $I$, and then $I$ can be exploited to address efficiently XAI queries for \emph{every} instance (forming the so-called varying part of the compilation problem).


%
%
%
%

\medskip

The main contribution of this paper is to show that \emph{rectification} \cite{DBLP:conf/ijcai/Coste-MarquisM21},
a belief change approach suited to the correction of binary classifiers, 
can be considered with profit for the distillation of a boosted tree $P$ into 
a decision tree $I$.  In a nutshell, rectifying $I$ by $P$ consists in modifying $I$ in a minimal way so that the resulting rectified tree classifies instances precisely as $P$ asks for.
 A valuable property of our rectification-based approach to distillation (and not shared by many other approaches to distillation, including distillation by retraining) is that \emph{it offers logical guarantees:} rectification ensures that the corrections that are targeted are effective.

In the general setting for rectification presented in \cite{DBLP:conf/ijcai/Coste-MarquisM21,coste-marquis23}, $P$ is any classification circuit and $I$ a formula. As a contribution, we show in the present paper how rectification can be specialized to the case when $P$ is a boosted tree and $I$ a decision tree, so that the rectification of $P$ by $I$ can be achieved incrementally and in a much more efficient way.
The principle of our approach is as follows: given a boosted tree $P$ and an initial decision tree $I$, for each instance $\vec x$ that is encountered at inference time, if $I(\vec x) \neq P(\vec x)$, an abductive explanation $t$ for $\vec x$ given $P$ \cite{IgnatievNM19} is computed using the approach presented in \cite{DBLP:conf/aistats/AudemardLMS23}. From it, a classification rule $R$ with premises $t$  
can be easily generated. 
By construction, this classification rule $R$ is deduced from $P$ in the sense that any instance $\vec x'$ covered by $t$ is necessarily classified by $P$ in the same way as $\vec x$. Then the current decision tree is rectified by $R$ to produce another decision tree and the process resumes. Notably, the rectification of $I$ by $R$ can be achieved in time polynomial in the size of $I$ plus the size of $R$. 


In our work, we also compared our rectification-based approach to distillation with a simple yet pure ML approach based on \emph{retraining} the model \cite{DBLP:journals/chinaf/Zhou19}. Basically, whenever a discrepancy between the prediction achieved by $I$ and by $P$ has been observed, the retraining approach consists in using $P$ as an oracle for updating the training set used to learn $I$ before learning $I$ again.
%
For each of the two approaches, we performed some experiments to assess the quality of the distillation produced, evaluated by measuring
the predictive performance of the corrected decision tree 
relative to $P$. 
The experimental results obtained have shown the interest of the distillation approach based on rectification compared to retraining.

Because the distillation of $P$ into $I$ may lead to a significant increase in the size of the decision tree (this increase being unavoidable in the worst case), it was also important to point out some empirical evidence to support the claim that the distillation of  $P$ into $I$ is computationally useful. 
To do so, we focused on a specific XAI query, namely computing a sufficient reason for a given instance $\vec x$. For this query, dedicated algorithms haven been implemented both for decision trees and for boosted trees, see in particular  \url{https://github.com/alexeyignatiev/xreason/} and \url{https://github.com/crillab/pyxai/} \cite{DBLP:journals/jair/IzzaIM22,DBLP:conf/aaai/IgnatievIS022,pyxai}. We took advantage of state-of-the-art algorithms for deriving sufficient reasons to compare the time needed to derive a sufficient reason for $\vec x$ from $I$ with the time needed to derive a sufficient reason for $\vec x$ from $P$, i.e., without distilling first $P$ into $I$. As soon as the former computation time is strictly smaller than the latter, the time spent in  distilling $P$ into $I$ can be balanced over sufficiently many instances $\vec x$. Our empirical results (based on average computation times and numbers of timeout for computing a sufficient reason) clearly show that distilling $P$ into $I$ is advantageous  despite the growth of $I$ it leads to.


\medskip
The datasets, the code used in our experiments and additional empirical results are available online at \cite{swh-dir-0689cda}. 

\section{Formal Preliminaries}

Let $X = \{x_1, \ldots, x_n\}$ be a set of Boolean variables and let $y$ be a Boolean variable not appearing in $X$. Literals $y$ and $\overline{y}$ are used to denote  the classes of positive and negative instances (respectively).
$X$ corresponds to the set of Boolean conditions appearing in the boosted tree $P$ and
it can be used to describe the instances \cite{Audemardetal23}. The set of all instances over $X$ is denoted by $\vec X$.
The elements of $X$ do not primarily represent independent conditions because they can come from the same numerical or categorical primitive 
attributes (for example, we can find in $P$ the condition $x_1 = (S > 30)$ relating to the numerical attribute $S$ but also the condition $x_2 = (S > 20)$ which is logically linked to it: $x_1$ cannot be true whereas $x_2$ would be false). A \emph{domain theory}, in the form of a logical formula $Th$ on $X$, specifies the links between non-independent Boolean conditions (for example, $Th = x_1 \Rightarrow x_2$).
Each instance  $\vec x$ of $\vec X$  can be considered as an interpretation on $X$ that satisfies $Th$. 
$\vec x$ is then viewed as a mapping associating each Boolean variable $x_i$ ($i \in [n]$) with $1$ if and only if the $i^{th}$ coordinate $\vec x_i$ of $\vec x$ is equal to $1$. 
This interpretation can be represented by a (canonical) term $t_{\vec x}$ on $X$, formed by the set (interpreted as a conjunction) of the positive literals $x_i$ ($i \in [n]$), such that $\vec x_i = 1$ and by the negative literals $\overline{x_i}$ ($i \in [n]$) such that $\vec x_i = 0$.

\begin{definition}
A \emph{binary classifier} on $X$ is a mapping $C$ from $\vec X$ to the set of Boolean values $\{0, 1\}$.
$\vec x \in \vec X$ is a positive instance if $C(\vec x) = 1$  and a negative one if $C(\vec x) = 0$.
\end{definition}
%

Any binary classifier can be represented by a classification circuit in the sense of \cite{DBLP:conf/ijcai/Coste-MarquisM21}:

\begin{definition}
A \emph{classification circuit} $\Sigma$ on $X \cup \{y\}$ is a circuit equivalent to a formula of the form $\Sigma_X \Leftrightarrow y$ where $\Sigma_X$ is a Boolean formula on $X$.
\end{definition}

Indeed, any binary classifier $C$ based on Boolean attributes $X$ (including decision trees and boosted trees) can be viewed as a Boolean formula $C_X$ on $X$ whose models are precisely the instances classified positively by $C$, i.e., satisfying $C(\vec x) = 1$. In general, there is no polynomial-time algorithm to get $C_X$ from $C$, but associated with $C$, we can always define the classification circuit $\Sigma = C_X \Leftrightarrow y$. 

\medskip
In the following, when $\Phi$ is a Boolean circuit or a formula on $X \cup \{y\}$ and $z$ is any variable from $X \cup \{y\}$, $\Phi(z)$ (resp. $\Phi(\overline{z})$) denotes the \emph{conditioning} of $\Phi$ by $z$ (resp. by $\overline{z}$).
$\Phi(z)$ (resp. $\Phi(\overline{z})$) is the circuit (or the formula) obtained by replacing in $\Phi$ any occurrence of $z$ by the Boolean constant $\top$ representing the truth value $1$ (resp. $\bot$ representing the truth value $0$).
When $\Sigma = \Sigma_X \Leftrightarrow y$ is a classification circuit on $X \cup \{y\}$, the set of models of $\Sigma(y)$ consists precisely of the models of $\Sigma_X$. Finally, when $\vec x \in \vec X$ is an instance, $\Phi(\vec x)$ denotes the iterative conditioning of $\Phi$ by each literal of $t_{\vec x}$. Thus, $\vec x \in \vec X$ is classified positively (resp. negatively) by 
$\Sigma$ when $\Sigma(\vec x)$ is equivalent to $y$ (resp. $\overline{y}$).

%
%
%
%
%
%
%

\medskip
Decision trees \cite{DBLP:books/wa/BreimanFOS84,DBLP:journals/ml/Quinlan86}  and boosted trees \cite{DBLP:journals/jcss/FreundS97} are two ML models that can be used to represent binary classifiers:

\begin{definition}
A \emph{decision tree} (resp. 
\emph{regression tree}\footnote{We use the term ``regression tree'' here simply because the leaves of such trees are labeled by real numbers and not by class identifiers. However, the task tackled in this paper is binary classification, not regression.}) $T$ on $X$ is a binary tree such that internal nodes are decision nodes labeled by elements $x$ of $X$ and leaves by Boolean constants (resp. real numbers). By convention, when following the left (resp. right) child of a decision node labelled by $x$, $x$ is set to false ($0$) (resp. true ($1$)). 
An instance $\vec x \in \vec X$ satisfies $T(\vec x) = v$ if and only if the unique path from the root of $T$ to a leaf which is compatible with $\vec x$  is a leaf labeled by $v$.
\end{definition}

\begin{definition}
%
A \emph {boosted tree} $P$ on $X$ is a set $\{T_1, \ldots, T_m\}$ of regression trees on $X$.
An instance $\vec x \in \vec X$ satisfies $F(\vec x) = 1$ iff $\sum_{i=1}^m T_i(\vec x) > 0$.
\end{definition}

\begin{figure}[t]
\centering
\scalebox{0.5}{
      \begin{tikzpicture}[scale=0.9, roundnode/.style={circle, draw=gray!60, fill=gray!5, very thick, minimum size=14.5mm, font=\LARGE},every node/.style={inner sep=1,outer sep=1}, squarednode/.style={rectangle, draw=red!60, fill=red!5, very thick, minimum size=5mm, font=\LARGE}]
        \node[roundnode](root) at (1.5,7){$S\!>\!30$};
        \node at (0,7){\LARGE $I$};
        \node[roundnode](n1) at (-1,5){$S\!>\!20$};
        \node[roundnode](n2) at (4,5){$R$};
        \node[squarednode](n11) at (-2,3){$0$};        
        \node[roundnode](n12) at (0,3){$PP$};   
        \node[squarednode](n121) at (-1,1){$0$};         
        \node[squarednode](n122) at (1,1){$1$};            
        \node[roundnode](n21) at (3,3){$PP$};
        \node[squarednode](n22) at (5,3){$1$};  
        \node[squarednode](n211) at (2,1){$0$};
        \node[squarednode](n212) at (4,1){$1$};   
        \draw[dashed] (root) -- (n1);
        \draw[dashed] (n1) -- (n11);
        \draw(n1) -- (n12);    
        \draw[dashed] (n12) -- (n121);
        \draw(n12) -- (n122);                
        \draw(root) -- (n2);
        \draw[dashed] (n2) -- (n21);
        \draw(n2) -- (n22);    
        \draw[dashed] (n21) -- (n211);
        \draw(n21) -- (n212);

        \node[roundnode](root2) at (7,7){$S\!>\!30$};
        \node at (5.5,7){\LARGE $T_1$};
        \node[squarednode](n2-1) at (6,5){$-1$};  
        \node[roundnode](n2-2) at (8,5){$R$};
        \node[squarednode](n2-21) at (7,3){$-1$};
        \node[squarednode](n2-22) at (9,3){$1$};  
        \draw[dashed] (root2) -- (n2-1);
        \draw(root2) -- (n2-2);
        \draw[dashed] (n2-2) -- (n2-21);
        \draw(n2-2) -- (n2-22);    
        
        \node[roundnode](root3) at (13,7){$S\!>\!30$};
        \node at (11.5,7){\LARGE $T_2$};
        \node[roundnode](n3-1) at (11.5,5){$S\!>\!20$};  
        \node[roundnode](n3-2) at (14.5,5){$PP$};  
        \node[squarednode](n3-11) at (10.5,3){$-2$};
        \node[squarednode](n3-12) at (12.5,3){$1$};   

        \node[squarednode](n3-21) at (13.5,3){$-1$};
        \node[squarednode](n3-22) at (15.5,3){$2$};     
        
        \draw[dashed] (root3) -- (n3-1);
        \draw(root3) -- (n3-2);
        \draw[dashed] (n3-1) -- (n3-11);
        \draw(n3-1) -- (n3-12);    
        \draw[dashed] (n3-2) -- (n3-21);
        \draw(n3-2) -- (n3-22); 

      \end{tikzpicture}
 }
 \newline
    \caption{
    A decision tree $I$ and a boosted tree $P$. $P$ is formed by two regression trees $T_1$ and $T_2$.
    For each tree, the dotted arc (resp. the solid arc) from a node labeled with a condition $\mathit{cond}$ corresponds to the assignment where $\mathit{cond}$ is false (resp. true). 
    For the sake of clarity, conditions $\mathit{cond}$ are expressed using the primitive attributes $S$, $R$ and $PP$ that have been used to learn $P$.\\
\label{fig:arbres}}
    \end{figure}
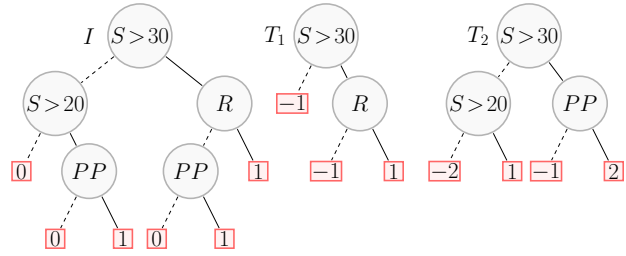

\begin{example}\label{ex:running}
As an illustrative example, let us consider a problem of credit allocation to bank customers. 
Each customer is characterized by an annual salary ($S$ a numerical attribute), a fact of having already reimbursed a previous loan ($R$ a Boolean attribute) and, whether or not, he has a permanent position ($PP$ a Boolean attribute).
Two binary classifiers are considered. On the one hand, a boosted tree $P$  consisting of two regression trees  ($T_1$ and $T_2$). 
On the other hand, a decision tree $I$. $P$ and $I$ are described in Figure \ref{fig:arbres}.
$X$ corresponds to the four Boolean conditions considered in this order $x_1 = (S > 30)$, $x_2 = (S > 20)$, $x_3 = R$, $x_4 = PP$.
Note that $x_1$ and $x_2$ are not independent. Indeed, an instance over $X = \{x_1, x_2, x_3, x_4\}$ is feasible only if it satisfies $Th = x_1 \Rightarrow x_2$.

We can easily check that $I$ is associated with a classification circuit on $X \cup \{y\}$ that is equivalent to $((x_1 \wedge x_3) \vee (x_2 \wedge x_4)) \Leftrightarrow y$ and that $P$ is associated with a classification circuit on $X \cup \{y\}$ that is equivalent to $(x_1 \wedge x_4) \Leftrightarrow y$.
We can also check that $I$ and $P$ classify the instances of $\vec X$ in the same way, except for the instances $(0, 1, 1, 1)$, $(0, 1, 0, 1)$ and $(1, 1, 1, 0)$. Indeed, these three instances are classified positively by $I$ and negatively by $P$.
\end{example}

A \emph{classification rule} is a rule that indicates thanks to its conclusion part (the right-hand side of the rule) how to classify any instance matching its premises part (the left-hand side of the rule).

\begin{definition}
%
A \emph{classification rule} over $y$ (resp. $\overline{y}$) is a formula of the form $R = \varphi_X \Rightarrow y$ (resp. $R = \varphi_X \Rightarrow \overline{y}$) where $\varphi_X$ is a formula on $X$.
\end{definition}

Classification rules do not state how to classify instances that are not covered by their left-hand side. For this reason, a classification rule (and more generally a set of such rules) does not represent a \emph{complete classifier}. Observe that when an instance $\vec x$ is covered by the left-hand side $\varphi_X$ of a rule $R$ (so that $R$ indicates how $\vec x$ must be classified), the iterative conditioning $R(\vec x)$ of $R$ by $t_{\vec x}$ is equivalent to the right-hand side of $R$. Thus, $R(\vec x)$ can be interpreted as the application of the (partial) classifier $R$ to instance $\vec x$, giving the corresponding class.
Furthermore, classification rules can be conflicting:
$R_1 = \varphi_X^1 \Rightarrow y$ and $R_2 = \varphi_X^2 \Rightarrow \overline{y}$ are said to be 
\emph{conflicting} if and only if  $Th \wedge \varphi_X^1 \wedge \varphi_X^2$ is consistent.
%
Given two conflicting classification rules $R_1 = \varphi_X^1 \Rightarrow y$ and $R_2 = \varphi_X^2 \Rightarrow \overline{y}$, one does not know how to classify an instance  $\vec x$ satisfying $\varphi_X^1 \wedge \varphi_X^2$ as the two rules give contradictory conclusions about the class of $\vec x$.

\medskip
Finally, one needs to make precise the notion of abductive explanation for an instance given a binary classifier \cite{IgnatievNM19}:

\begin{definition}
%
An \emph{abductive explanation} for $\vec x \in \vec X$ given a binary classifier $C$ on $X$ is a term $t$ on $X$ such that $t$ covers $\vec x$ (i.e., $t \subseteq t_{\vec x}$) and for all $\vec x' \in \vec X$ such that  $t$ covers  $\vec x'$, one has $C(\vec x') = C(\vec x)$.
\end{definition}

Such an abductive explanation $t$ provides a set of conditions (literals) corresponding to characteristics of the input instance $\vec x$ and explaining why the instance $\vec x$ is classified by $C$ in the way it has been classified.

\begin{example}[Example \ref{ex:running}, cont'ed]
$t= \overline{x_1} = \overline{(S > 30)}$ is an abductive explanation for  $(0, 1, 1, 1)$ given  $P$. 
The fact that the salary of the incomer is less than or equal to 30k\$ is sufficient to explain why, according to $P$, the loan requested should not be granted. 
\end{example}

\section{Rectifying a Classification Circuit}

In the general case, when a classification circuit $\Sigma = \Sigma_X \Leftrightarrow y$ is rectified by a  formula $F$ on $X \cup \{y\}$, the result of the rectification process \cite{coste-marquis23} is the classification circuit $\Sigma \star F$ defined by 
$$\Sigma \star F = \Sigma_X^F \Leftrightarrow y, \mbox{~where}$$ $$\Sigma_X^F \equiv (\Sigma_X \wedge \neg (F(\overline{y}) \wedge \neg F(y))) \vee (F(y) \wedge \neg F(\overline{ y})).$$

$\Sigma^F_X$ characterizes the instances to be classified as positive after the rectification of $\Sigma$ by $F$. Those instances can be gathered into
two sets: the instances that are consistently asked to be classified as positive by $F$ (they are characterized by the subformula
$F(y) \wedge \neg F(\overline{y})$ of $\Sigma^F_X$), and the instances that were already classified as positive by $\Sigma$ provided that
$F$ did not consistently ask them to be classified as negative (those instances are characterized by the subformula
$\Sigma_X \wedge \neg(F(\overline{y}) \wedge \neg F(y))$ of $\Sigma^F_X$).

\begin{example}[Example \ref{ex:running}, cont'ed]
Suppose that the rectification formula $F$ is the classification rule $F = \overline{x_4} \Rightarrow \overline{y}$. This classification rule can be deduced from the classification circuit $(x_1 \wedge x_4) \Leftrightarrow y$ associated with $P$ ($F$ is a logical consequence of $(x_1 \wedge x_4) \Leftrightarrow y$). 
We have $F(y) \equiv x_4$ and $F(\overline{y}) \equiv \top$, so that $F(y) \wedge \neg F(\overline{y}) \equiv \bot$ and $\neg(F(\overline{y}) \wedge \neg F(y)) \equiv x_4$. Thus, rectifying the classification circuit $((x_1 \wedge x_3) \vee (x_2 \wedge x_4)) \Leftrightarrow y$ associated with $I$ by the formula $F$ leads to a classification circuit that is equivalent to
$(((x_1 \wedge x_3) \vee (x_2 \wedge x_4)) \wedge x_4) \Leftrightarrow y$, thus equivalent to $(((x_1 \wedge x_3) \vee x_2) \wedge x_4) \Leftrightarrow y$.
\end{example}

Interestingly, in the specific context considered in this paper, i.e., $\Sigma_X$ is a decision tree on $X$ and $F = P_X \Leftrightarrow y$ where $P$ is a boosted tree, the previous definition of $\Sigma \star F$ can be simplified significantly, leading to improved computations. 
The next three propositions are the key results on which our distillation method is based. Proposition \ref{prop:simpl} shows that the general definition of a rectified classification circuit can be simplified when the rectification formula is a classification rule. Proposition \ref{prop:all} shows that the rectification of a classification circuit by a conjunction of classification rules can be achieved on a rule-per-rule basis.
Finally, Proposition \ref{prop:decision} indicates how classification rules can be generated from abductive explanations.



\begin{proposition}\label{prop:simpl}
Let  $\Sigma = \Sigma_X \Leftrightarrow y$ be a classification circuit on  $X \cup \{y\}$. 
Let  $R = \varphi_X \Rightarrow y$ (resp. $R = \varphi_X \Rightarrow \overline{y}$) be a classification rule over $y$  (resp.  $\overline{y}$).
We have $\Sigma \star R \equiv (\Sigma_X \vee \varphi_X) \Leftrightarrow y$ (resp. $\Sigma \star R \equiv (\Sigma_X \wedge \neg \varphi_X) \Leftrightarrow y$).
\end{proposition}

The previous example illustrates this proposition.

The next result shows that rectifying a classification circuit $\Sigma$ by a (conjunctively-interpreted) \emph{set $F$ of classification rules deduced from another classification circuit} amounts to rectify $\Sigma$ by each rule from $F$ in an iterative fashion (the order according to which the rules are considered does not matter because those rules are never conflicting):
%
%

\begin{proposition}\label{prop:all}
Let  $\Sigma = \Sigma_X \Leftrightarrow y$ be a classification  circuit on  $X \cup \{y\}$.
Let  $\Phi = \Phi_X \Leftrightarrow y$ be another classification circuit on $X \cup \{y\}$. 
Let  $\{R_1, \ldots, R_k\}$ be a set of classification rules that can be deduced from $\Phi$.
We have  $$\Sigma \star (R_1 \wedge \ldots \wedge R_k) \equiv (\Sigma \star R_1) \star \ldots \star R_k.$$
\end{proposition}



Finally, the next proposition shows how to deduce classification rules from a circuit $\Phi_X \Leftrightarrow y$, using abductive explanations for instances given $\Phi_X$: 

\begin{proposition}\label{prop:decision}
Let $C_X \Leftrightarrow y$ be a classification circuit on $X \cup \{y\}$ associated with a binary classifier $C$. 
Let  $\vec x \in \vec X$ be an instance such that $C(\vec x) = 1$ (resp. $C(\vec x) = 0$).
Let $t$ be an abductive explanation for an instance $\vec x \in \vec X$ given $C$. 
$R = t \Rightarrow y$ (resp. $R = t \Rightarrow \overline{y}$)
is a classification rule over $y$ (resp. $\overline{y}$) 
that is implied by $C_X \Leftrightarrow y$.
\end{proposition}

\section{Distilling Boosted Trees into Decision Trees}\label{sec:distilling}

By combining Propositions \ref{prop:simpl}, \ref{prop:all} and \ref{prop:decision}, one can define an incremental (and possibly partial) distillation process of boosted trees $P$ into decision trees $I$. The idea is to consider only instances  $\vec x$ that are misclassified by $I$ (i.e., those classified differently by $P$) 
as soon as they appear at inference time, and whenever such an instance $\vec x$ is encountered, to correct the classification circuit $I_X \Leftrightarrow y$ associated with $I$ by a classification rule $R$ that covers $\vec x$ and is implied by the classification circuit $P_X \Leftrightarrow y$ associated with $P$. At each correction step, the set $\vec X_I^{\pm} = \{\vec x \in \vec X \mid I(\vec x) \neq P(\vec x)\}$ of instances from $\vec X$ that are, according to $P$, misclassified by $I$
is reduced.

The choice for such a \emph{lazy but opportunistic} approach to distillation comes from spatial complexity results showing that the
full rectification of $I_X \Leftrightarrow y$ by $P_X \Leftrightarrow y$ in one step would be  out of reach in the worst case. If the classification circuit $P_X \Leftrightarrow y$ can be characterized by an equivalent conjunction of non-conflicting classification rules, given by
$\bigwedge_{\vec x : P(\vec x) = 1} (t_{\vec x} \Rightarrow y) \wedge \bigwedge_{\vec x : P(\vec x) = 0 } (t_{\vec x} \Rightarrow \overline{y})$. this conjunction (a \cnf\ formula over $X \cup \{y\}$) is not directly usable, as it is exponentially large (it contains as many rules as instances). Even if more compact representations of this set of rules as a \cnf\ formula exist in general, any \cnf\ formula  equivalent to $P_X$
is exponential in the size of $P$ in the worst case \cite{DBLP:conf/ijcai/ColnetM23}.
Moreover, any decision tree equivalent to such a \cnf\ formula would be also, in the worst case, exponential in the size of the \cnf\ formula \cite{DBLP:conf/aaai/AudemardBBKLM22}. 
Algorithm \ref{algo} makes precise how one step of the incremental distillation process of $P$ into $I$ (the step triggered by $\vec x$) is achieved. An abductive explanation $t$ for $\vec x$ given $P$ is first computed. A classification rule $R$ that is implied $P_X \Leftrightarrow y$
(cf. Proposition \ref{prop:decision}) is formed using $t$. Then, using Proposition \ref{prop:simpl}, the classification circuit $\Sigma_X \Leftrightarrow y$ where $\Sigma_X = I$ is rectified by $R$. In detail, a decision tree $I^R$ equivalent to $\Sigma_X \vee \varphi_X$ (or to
$\Sigma_X \wedge \neg \varphi_X$ depending on the right-hand side of $R$) can be generated efficiently by looking at the root-to-leaf paths $p$ of $\Sigma_X$ and \emph{rectifying each of them separately, in parallel}.\footnote{This approach can be easily extended to the multi-class and even to the multi-label classification setting.} Basically, every root-to-leaf path $p$ of a decision tree can be associated with a term $t$, which can be defined by induction as follows: let $t$ be the empty term; if $p$ reduces to a leaf node, then return $t$; otherwise, $p$ starts with a decision node (the root of the tree) which is labelled by a Boolean variable $x$; then add to $t$ literal $x$ (resp. literal $\overline{x}$) if $p$ goes right (resp. left) from the decision node and resume from the child node that has been reached. When rectifying $\Sigma_X$ by $R$, only those paths $p$ of $\Sigma_X$ associated with terms $t$ such that $t \wedge \varphi_X \wedge Th$ is consistent need to be considered. More precisely, among them only those paths leading to a $0$-leaf (resp. a $1$-leaf) need to be updated when the right-hand side of $R$ is $y$ (resp. $\overline{y}$). Thus, when the right-hand side of $R$ is $y$ (resp. $\overline{y}$) updating $p$ simply consists in replacing its $0$-leaf node (resp. its $1$-leaf node) by a decision tree representing the conjunction of the literals from $t \setminus \varphi_X$ (resp. the negation of the conjunction of the literals from $t \setminus \varphi_X$). Finally, the domain theory $Th$ associated with $I$ can be leveraged to simplify the resulting tree $I^R$. In a bottom-up way, starting from the leaf of a branch $p$ of $I^R$ up to the root of $I^R$, an arc of $p$ can be removed when the literal $\ell$ labelling it is a logical consequence of $(p \setminus \{\ell\}) \wedge Th$. Furthermore, any internal node of $I^R$ with a left subtree identical to its right subtree can be replaced by one of its two subtrees (see \cite{coste-marquis23} for details). This simplification step is fundamental in practice to limit the growth of the tree.

\begin{algorithm}[H]
  \caption{The incremental distillation of a boosted tree into a decision tree. }\label{algo} 
  \begin{algorithmic}
  \Require{a decision tree $I$ and a boosted tree $P$ over $X$, an instance $\vec x \in \vec X$ such that $\vec x \in \vec X_I^{\pm}$.}\;
  \Ensure{a decision tree $I^R$ such that $\vec X_{I^R}^{\pm} \subset \vec X_{I}^{\pm}$.}\;
  \State $t \leftarrow abductive-expl(P, \vec x)$\; 
  \If{$P(\vec x) = 1$}
  \State $R \leftarrow (t \Rightarrow y)$
  \Else
  \State $R \leftarrow (t \Rightarrow \overline{y})$\;
  \EndIf\;
  \State $I^R \leftarrow rectify(I, R)$\;
  \State $I^R \leftarrow simplify(I^R)$\;\\
  \Return($I^R$)\;
\end{algorithmic}
\end{algorithm}

\begin{example}[Example \ref{ex:running} cont'ed]
Consider the instance $\vec x = (0, 1,1, 1)$. 
As  $I(\vec x) = 1$ and $P(\vec x) = 0$, we have $\vec x \in \vec X_{I}^{\pm}$, so one needs to correct the misclassification done by $I$. Using the approach introduced in \cite{DBLP:conf/aistats/AudemardLMS23}, the abductive explanation  $t = \overline{x_1} = \overline{(S > 30)}$ for $\vec x$ given  $P$ is first computed. 
Since  $P(\vec x) = 0$, from Proposition \ref{prop:decision}, we know that the classification rule  $R = \overline{x_1} \Rightarrow \overline{y}$ is implied by $P_X \Leftrightarrow y$.

Then, one rectifies the classification circuit $\Sigma = I_X \Leftrightarrow y$ by the rule $R$ and produces a circuit $\Sigma \star R$ which is equivalent to  $(I_X \wedge \neg t) \Leftrightarrow y$, following Proposition \ref{prop:simpl}.
To do it, we generate a decision tree $I^R$ such that $I^R_X$ is equivalent to  $I_X \wedge \neg t$. $I^R$ is obtained by updating every path of $I$ corresponding to a term that is consistent with the premises $t = \overline{x_1} = \overline{(S > 30)}$ of $R$ but with a $1$-leaf (since the conclusion $\overline{y}$ of $R$ asks for a $0$-leaf). A single path of $I$ needs to be updated, the one associated with the term $\overline{(S > 30)} \wedge (S > 20) \wedge PP$ (see Figure \ref{fig:arbres}). Updating it simply consists in replacing its $1$-leaf by a $0$-leaf, resulting is the tree shown in Figure \ref{fig:arbrerectifie} (left sub-figure), in which the replaced leaf appears in red. 
This tree can then be simplified (the resulting tree is shown in the sub-figure on the right). 


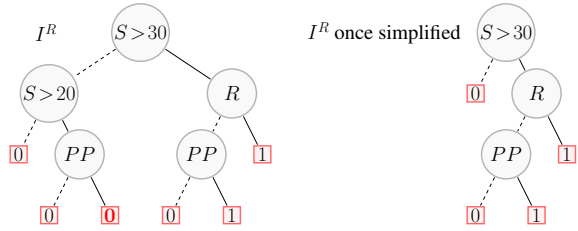
\begin{figure}[t]
\centering
\scalebox{0.45}{
      \begin{tikzpicture}[scale=0.9, roundnode/.style={circle, draw=gray!60, fill=gray!5, very thick, minimum size=14.5mm, font=\LARGE},every node/.style={inner sep=1,outer sep=1},
      squarednode/.style={rectangle, draw=red!60, fill=red!5, very thick, minimum size=5mm, font=\LARGE}]
        \node[roundnode](root) at (0,7){$S\!>\!30$};
        \node at (-3.1,7){\LARGE $I^R$}; 
        \node[roundnode](n1) at (-3,5){$S\!>\!20$};
        \node[roundnode](n2) at (3,5){$R$};
        \node[squarednode](n11) at (-4,3){$0$};        
        \node[roundnode](n12) at (-2,3){$PP$};   
        \node[squarednode](n121) at (-3,1){$0$};    
        \node[squarednode](n122) at (-1,1){$\textcolor{red}{\mathbf{0}}$};
                     
        \node[roundnode](n21) at (2,3){$PP$};
        \node[squarednode](n22) at (4,3){$1$};  
                
        \node[squarednode](n211) at (1,1){$0$};
        \node[squarednode](n212) at (3,1){$1$};

        \draw[dashed] (root) -- (n1);
        \draw[dashed] (n1) -- (n11);
        \draw(n1) -- (n12);    
        \draw[dashed] (n12) -- (n121);
        \draw(n12) -- (n122);                
        \draw(root) -- (n2);
        \draw[dashed] (n2) -- (n21);
        \draw(n2) -- (n22);    
        \draw[dashed] (n21) -- (n211);
        \draw(n21) -- (n212);   
                       
        \node[roundnode](root2) at (12,7){$S\!>\!30$};
        \node at (8,7){\LARGE $I^R \mbox{~once simplified}$};
        \node[squarednode](n2-1) at (11,5){$0$};  
        \node[roundnode](n2-2) at (13,5){$R$};
        \node[roundnode](n2-21) at (12,3){$PP$};
        \node[squarednode](n2-211) at (11,1){$0$}; 
        \node[squarednode](n2-212) at (13,1){$1$};  
        \node[squarednode](n2-22) at (14,3){$1$};
        \draw[dashed] (root2) -- (n2-1);
        \draw(root2) -- (n2-2);
        \draw[dashed] (n2-2) -- (n2-21);
        \draw(n2-2) -- (n2-22);    
        \draw[dashed] (n2-21) -- (n2-211);
        \draw(n2-21) -- (n2-212); 

      \end{tikzpicture}
 }
 \newline
    \caption{A decision tree $I^R$ such that $I^R_X \Leftrightarrow y$ is equivalent to $(I_X \Leftrightarrow y) \star R$ (left). 
    An equivalent decision tree, obtained by simplification (right). 
    \label{fig:arbrerectifie}}
    \end{figure}
Note that the correction step achieved by rectifying $I_X \Leftrightarrow y$ by $R$ also corrects the classification error made by $I \Leftrightarrow y$ on the  instance $(0, 1, 0, 1)$. Indeed, the decision tree $I^R$ classifies all instances in the same way as $P$, except for $(1, 1, 1, 0)$, which will eventually be corrected if the instance $(1, 1, 1, 0)$ is considered in the future.
\end{example}

In our approach, each correction step is thus triggered by \emph{a single instance} $\vec x$. Considering a population instead (i.e., a batch of instances classified in the same way by $P$) would not be possible in general because two instances can be classified in the same way by $P$ for incompatible reasons (i.e., the two sets of abductive explanations can be disjoint). Thus, as a matter of illustration, consider the running example again: the instances $(1, 1, 1, 0)$ and $(0, 1, 1, 1)$ are classified negatively by $P$ but they do not share any common abductive explanation (especially, $t = x_2 \wedge x_3$ is not such an explanation since it also covers the instance $(1, 1, 1, 1)$ which is classified positively by $P$).

\medskip
Notably, our rectification-based approach ensures that $I$ is \emph{made logically closer} to $P$ at each correction step (i.e., the number of instances $\vec x$ such that $I(\vec x) = P(\vec x)$ increases at each step). Stated differently, Algorithm \ref{algo} is correct w.r.t. its specification:

\begin{proposition}\label{prop:converge}
The decision tree $I^R$ computed by Algorithm \ref{algo} from a decision tree $I$, a boosted tree $P$, and an instance $\vec x \in X_I^{\pm}$ is such that $X_{I^R}^{\pm} \subset X_I^{\pm}$.
\end{proposition}

This guarantee is offered whatever the decision tree $I$ one starts with, especially in the restricted case when the initial tree $I$ consists of a single node (e.g., a leaf labelled with the majority class) or when $I$ already is at start equivalent to $P$.
Actually, the choice of the initial decision tree $I$ impacts only the number of correction steps that will be needed to achieve a full distillation.

The guarantee that $X_{I^R}^{\pm} \subset X_I^{\pm}$ is also ensured whatever the rule $R$ that is computed by Algorithm \ref{algo}. When forming classification rules $R$ from abductive explanations $t$, subset-minimal abductive explanations appear as the best candidates since they lead to the most general (i.e., logically strongest) classification rules $R$. Indeed, since such rules cover more instances than rules that are less general, using them may lead to diminishing the number of correction steps of the distillation process. Thus, ideally, they should be preferred. However, deriving subset-minimal abductive explanations given boosted trees is intractable \cite{DBLP:conf/aistats/AudemardLMS23}. This explains why in our implementation we focused on tree-specific explanations. As shown in \cite{DBLP:conf/aistats/AudemardLMS23}, such abductive explanations may contain (arbitrarily) many redundant characteristics (i.e., they are not subset-minimal in general) but they can be computed in polynomial time. 

%
%
The order with which rectifications have been made has also no impact on the resulting tree once all the misclassified instances $\vec x$ have been handled, provided that the same rule has been extracted whatever the step at which $\vec x$ is considered (this is a direct consequence of Proposition \ref{prop:all}). Indeed, in this case, the resulting tree represents the same binary classifier whatever the order with which the misclassified instances have been encountered.

On the contrary, at each step, the misclassified instance $\vec x$ that is encountered has a big impact on the abductive explanation $t$ that is computed (since $t$ must be an abductive explanation for $\vec x$), thus on the classification rule $R$ that is derived from this explanation, and, as a consequence, on the misclassified instances that are covered by this rule. Similarly, the abductive explanation $t$ for $\vec x$ given $P$ that is chosen to form $R$ matters.\footnote{An instance $\vec x$ may have exponentially many subset-minimal abductive explanations given a binary classifier $C$, especially when $C$ is a boosted tree. It may also have exponentially many tree-specific explanations.}
Indeed, the number of instances that remain misclassified after a preset number $k$ of rectification steps can greatly vary depending on the instance considered  at each step and its chosen explanation. Especially, if the correction process is interrupted before exhaustion (i.e., the distillation of $P$ into $I$ is partial), significantly different decision trees can be generated after $k$ rectification steps depending on the choices made.

\section{Experiments}

\subsection{Empirical protocol} 

In our experiments, we considered various datasets for binary classification, coming  from two well-known open repositories:
UCI (\url{https://archive.ics.uci.edu/ml/index.php}) and openML (\url{https://www.openml.org/}).
Instances of these datasets contain attributes of various types (numerical, categorical or Boolean). Each instance is associated with a class $c$, which is equal to $1$ or to $0$ depending on whether the instance is positive or negative.

 \begin{table}[t]
 \caption{Description of dataset and accuracy before distillation.}\label{description}
 \renewcommand{\arraystretch}{1.2}
 \setlength{\tabcolsep}{6pt}
\centering
 \resizebox{\linewidth}{!}{\Large
 \begin{tabular}{ l r r r  r r  r r r r}
 \toprule
 Dataset & $|E|$ & $|F|$ & $|B|$ & $\%I_o$ & $\%I_d$ & $\%P$ & Repository & $|T_{I_o}^{\pm}|$ & $|T_{I_d}^{\pm}|$\\
 \midrule
 {\tt bank} & 4521 & 48 & 521  & 87.50\% &  85.46\%  & 88.0\% & UCI & 125 & 153\\
 {\tt biodegradation} & 1054 & 41 & 242  &  81.75\%  & 81.08\% & 84.68\% & openML & 42 &48 \\
 {\tt australian} & 690 & 38 & 174  & 81.63\%  & 80.39\% & 86.20\% & openML & 26 & 32\\
 {\tt bupa} & 345 & 5 & 108  & 89.58\%  & 89.58\% & 97.26\% & UCI & 8 & 8\\
 {\tt german} & 1000 & 58 & 105  & 94.28\% & 93.57\% & 95.71\% & UCI & 13 & 20\\
 {\tt contraceptive} & 1473 & 21  & 68  & 68.44\%  &  62.13\% & 73.87\% & UCI & 77 & 126\\
 {\tt cleveland} & 303 & 23 & 42  & 78.57\%  & 76.19\%  & 87.5\% & openML & 12 & 13\\
 {\tt compas} & 6172 & 11 & 33  & 68.28\%  & 65.85\% & 69.29\% & openML & 125 & 254  \\
 {\tt cnae} & 1080 & 856 & 15  & 96.13\%  & 95.97\% &  96.47\% & UCI & 5 & 5  \\
 {\tt breast-tumor} & 286 & 37 & 58  & 52.7\%  & 52.5\% &  53.5\% & UCI & 29 & 30  \\
 {\tt balance\_0\_vs\_1} & 625 & 4 & 17  & 88.37\%  &84.88\% &  90.90\% & UCI & 6 & 18  \\
 {\tt balance\_0\_vs\_2} & 625 & 4 & 17  & 82.97\%  &82.60\% &  90.14\% & UCI & 7 & 8  \\
  {\tt balance\_1\_vs\_2} & 625 & 4 & 17  & 88.90\%  &80.00\% &  94.21\% & UCI & 16 & 31  \\
 \bottomrule
\end{tabular}}
 \end{table}

Each dataset has been partitioned into two subsets, the first one (70\% of the available instances) was used for training and the second one (30\% of instances) was used to trigger the corrections and for the testing purpose (i.e., to measure the predictive performance of the classifiers). 
 

One started by learning a precise model $P$, here, a boosted tree. %
For each dataset, $P$ has been learned using the algorithm provided in the XGBoost library \cite{Chen16}, adjusting hyperparameter values using grid search in an attempt to achieve predictive performance that is good enough. 
The hyperparameterization of $P$ (i.e., adjusting the values of max depth, n\_estimators, and learning rate) was based only on the training set. Table \ref{tab:best-hyperparams} indicates the values of the hyperparameters that have been used for learning $P$, as well as the number of nodes in $P$.

\begin{table}[t]
\caption{Best hyperparameters found for $P$ for each dataset.}
\label{tab:best-hyperparams}
 \renewcommand{\arraystretch}{1.2}
 \setlength{\tabcolsep}{6pt}
\centering
\small  
 \resizebox{.80\linewidth}{!}{\Large
\begin{tabular}{lcccc}
\hline
{Dataset} & {Learning Rate} & {Max Depth} & {Number of Estimators} & {Number of Nodes} \\
\hline
 {\tt bank} & 0.2 & 4 & 200 & 2112  \\
 {\tt biodegradation} & 0.1 & 9 & 200 & 2222 \\
 {\tt australian} & 0.1 & 8 & 100 & 1520 \\
 {\tt bupa} & 0.02 & 6 & 200 & 984 \\
 {\tt german} & 0.2 & 5 & 100 & 994 \\
 {\tt contraceptive} & 0.2 & 6 & 150 & 4168 \\
 {\tt cleveland} & 0.02 & 7 & 150 & 1016\\
 {\tt compas} & 0.02 & 6 & 100 & 4022\\
 {\tt cnae} & 0.1 & 3 & 100 & 588\\
 {\tt breast-tumor} & 0.3 & 7 & 200 & 2188 \\
 {\tt balance\_0\_vs\_1} & 0.1 & 6 & 100 & 1598 \\
 {\tt balance\_0\_vs\_2} & 0.1 & 6 & 100 & 1304\\
 {\tt balance\_1\_vs\_2} & 0.1 & 6 & 100 & 1232 \\
\hline
\end{tabular}}
\end{table}
Once $P$ has been learned, every instance of each dataset has been translated into an instance represented in the space of the $n$ Boolean conditions used in $P$ (thus, the resulting instances are described solely using Boolean attributes). $L$ (resp. $T$) denotes the resulting set of instances used to learn decision trees (resp. to trigger corrections and to test the predictive performance of the trees).

Then, decision trees $I$ have been learned from $L$ using the algorithm provided in the scikit-learn library \cite{scikit-learn}. Two configurations have been tested: one for which hyperparameters have been set to their \emph{default values}
and one for which hyperparameters have been \emph{optimized}. 

The hyperparameterization of $I$ (here, adjusting the value of max depth) was based on the training set $L$.
In the default configuration, the depth of $I$ is not bounded a priori (whatever the step): any internal node
$N$ of $I$ is decomposed whenever the subset of the training set verifying all
the conditions of the path going from the root of the tree to $N$ only contains instances of
the same class (the node $N$ is said to be pure).
In the optimized configuration, the depth of $I$ has been tuned in order to achieve a better predictive performance and avoid overfitting. It has been set, for each dataset, to the value given in Table \ref{tab:resultats}, step $0$, column $D$. 
Whatever the step, the depth of retrained trees has been limited to this value. 

The default configuration typically leads to constructing decision trees that overfit, offering in general a lower accuracy when assessed on $T$. Nevertheless, it makes sense to consider this default configuration for two reasons.
On the one hand, because of the limited precision of the trees considered initially under this configuration, the correction steps to be carried out can be numerous. On the other hand, 
it is particularly favorable for ensuring correction guarantees, even when retraining is used for the correction purpose. 
Indeed, in practice, the choices made at each decision node under the default configuration ensure 
that the instances of the training set are associated with their 
expected class (the one given in the training set). So, every time an instance is added to the training set so as to correct its classification, the new decision tree learned after this addition 
classifies the instance correctly: the desired correction is thus achieved. 
For each configuration, a repeated random sub-sampling cross validation process
has been achieved: we learned $10$ decision trees $I$ from $L$, retaining 70\% of instances from $L$ for training each tree. The median accuracy $\%I_o$ (resp. $\%I_d$) obtained for the 10 decision trees has been measured on the corresponding test set $T$ for optimized (resp. default) configuration.

The next step was to correct the decision tree $I$ at hand whenever necessary and whatever the configuration used to learn it. 
%
For each $(\vec x, c)$ in $T$, $P(\vec x)$ is considered as the "true" class of $\vec x$. $T_I^{\pm} = \{(\vec x, c) \in T : I(\vec x) \neq P(\vec x)\}$ is the subset of instances in $T$ that, according to $P$, are not classified correctly by $I$. The accuracy $I_P$ of $I$ \emph{relative to $P$}, empirically measured on $T$, is given by $1 - \frac{|T_I^{\pm}|}{|T|}$.
So, if $I(\vec x) = P(\vec x)$ for every $(\vec x, c) \in T$, $I_P$ is $100\%$. Thus, $1 - I_P$ indicates the proportion of instances in $T$ that still need correction. By the way,  please keep in mind that the value of $I_P$ only indicates the extent to which $I$ classifies instances in the same way as $P$, but does not give any information about the actual predictive performance of $P$.

For each $\vec x \in T_I^{\pm}$, we computed an abductive explanation (to be more precise, a tree-specific explanation) $t$ for $\vec x$ given $P$ using the code furnished in the PyXAI library \cite{pyxai}. This explanation $t$ gives rise to the classification rule $R = t \Rightarrow y$ when $P(\vec x) = 1$ and to the classification rule $R = t \Rightarrow \overline{y}$ when $P(\vec x) = 0$, which indicates a reason (namely, $t$) for the classification achieved by $P$ for every instance covered by $t$. Then:

\begin{itemize}
\item As to the retraining approach, 
at each step, a small sample of classified instances $(\vec x', P(\vec x))$ containing $(\vec x, P(\vec x))$ and such that $t$ covers $\vec x'$ is added to $L$. More precisely, a limited ratio $r$ ($r = 1\%$ in the experiments) of the total number of instances that can be produced using the $n$ Boolean conditions occurring in $P$ is generated and the number of instances in the sample is limited to a preset bound $b$ (equal to $100$ in the experiments). 
This restriction is necessary to prevent an unmanageable growth of the training set $L$ at each step since the number of instances covered by $t$ is exponential in $n - |t|$. 
The sample is obtained by randomly choosing, according to a uniform distribution, conditions from $P$ not present in $t$ until $\mathit{max}(r \times 2^{n-|t|}, b)$ instances have been generated. Only those instances that are feasible given the underlying domain theory $Th$ \cite{DBLP:conf/aaai/GorjiR22} are retained. 
Then, every instance $(\vec x', c)$ from the resulting training set such that 
the premises $t$ of $R$ covers $\vec x'$ and the conclusion of $R$ is different of $c$ 
is removed from the training set. 
Finally, a new training of the model $I$ using the updated training set is achieved. 
\item As to the rectification approach, the current decision tree $I$ is corrected with the classification rule $R$, resulting in a new decision tree.
\end{itemize}

After each correction, $T_I^{\pm}$ has been recalculated before moving on to the next correction (this is necessary as $I$ has been modified).
\medskip

For each of the two correction approaches and for each of the two configurations tested, we measured after each correction (for each resulting decision tree $I$):
\begin{itemize}
\item The accuracy $I_P$ of the decision tree $I$ relative to $P$ (this accuracy is estimated on the test set $T$).
\item The size of the decision tree $I$ (the number $N$ of its nodes) and the depth $D$ of $I$.
\end{itemize}



In order to be sure that computational benefits may result from the distillation process, it was also important to check that the time spent in distilling $P$ into a decision tree can be balanced. We made some experiments to test whether this is actually the case: using the rectification approach, $P$ has been distilled into a decision tree in an incremental way, up to the step $f$ from which $I_P = 100\%$ (i.e., when the resulting  tree classifies all the instances of $T$ as the boosted tree $P$) and the overall compilation / distillation time required to reach such a tree $I$ has been measured. This has been achieved for each of the $10$ decision trees learned from $L$ at start, using the default configuration (i.e., the depth of the initial tree was not bounded a priori). 
Then, for each dataset, 100 instances $\vec x$ were picked up uniformly at random. A sufficient reason for each $\vec x$ given $I$ has been computed for each $I$ using a deletion-based algorithm \cite{IgnatievNM19} and a sufficient reason for each $\vec x$ given $P$ has been computed using the algorithm put forward in \cite{DBLP:conf/aaai/IgnatievIS022}.
For the two algorithms, a timeout of 180 seconds was considered per instance. 

\medskip
Datasets used in the experiments are described in Table \ref{description}. 
We kept in the experiments only those datasets from UCI and openML leading to boosted trees $P$ achieving higher accuracies than the corresponding decision trees $I$ considered at start, i.e., at step $0$ (in the remaining case, the interest to distill $P$ into another decision tree than $I$ is dubious).
From left to right, the table indicates the number $|E|$ of examples (instances) in the dataset, the number $|F|$ of features (attributes) used to  describe the instances initially, the number $|B|$ of Boolean conditions used in the dataset once binarized using $P$, the median accuracy  $\%I_o$ (resp. $\%I_d$) obtained for the 10 decision trees considered under the optimized (resp. default) configuration, the accuracy $\%P$ of the boosted tree, the repository from which the dataset comes from, and finally the median number $|T_{I_o}^{\pm}|$ (resp. $|T_{I_d}^{\pm}|$) of instances of the  test set $T$ classified differently by $I_o$ (resp. $I_d$) and $P$. We can check from the table that, as expected, the median accuracy of the decision trees learned under the default configuration typically is lower than the median accuracy of the decision trees learned under the optimized configuration, and the median number of instances from $T$ to be corrected for decision trees learned under the default configuration never is at least as large as the median number of instances from $T$ to be corrected for decision trees learned under the optimized configuration.

All the experiments have been conducted on computers with 2 quad-core Intel(R) Xeon(R) CPU E5-2643 0 @ 3.30GHz, each equipped with 32GiB of memory.
%

\subsection{Empirical results}

\renewcommand{\arraystretch}{1.2}
\setlength{\tabcolsep}{2.5pt}
\begin{table}[H]
\centering        
\resizebox{\linewidth}{!}{\Large
\begin{tabular}{l rr rr r rr r rr r rr r rr r r rrr }
    \toprule
    \multicolumn{3}{c}{} & \multicolumn{16}{c}{Correction steps (default configuration)} \\
    \cmidrule{4-20}

     \multicolumn{2}{c}{Dataset} &&\multicolumn{3}{c}{0} &&\multicolumn{3}{c}{1}&&\multicolumn{3}{c}{2}&&&\multicolumn{4}{c}{Final step (rec)}  \\
    \cmidrule{4-6} \cmidrule{8-10} \cmidrule{12-14} \cmidrule{17-20}
    \multicolumn{3}{c}{} & $I_P$&$N$&$D$ &&   $I_P$&$N$&$D$   && $I_P$&$N$&$D$  && &$I_P$&$N$&$D$&$f$  \\ 
    \cmidrule{1-2}  \cmidrule{4-6} \cmidrule{8-10} \cmidrule{12-14} \cmidrule{17-20}
    \multirow{2}{*}{\texttt{bank}} & rec && 88.76 & 512 & 22 && 88.83 &  562 & 24 &&  88.90 & 602 & 27 & & & 100.0 & 12273 & 39 & 153\\
                  & ret &&  88.76   & 512 & 22 && 89.60 &  520 & 23 &&  89.75 & 521 & 22 & & &  94.17 & 941 & 34&-\\
    \cmidrule{1-2} \cmidrule{4-6} \cmidrule{8-10} \cmidrule{12-14} \cmidrule{17-20}

    \multirow{2}{*}{\texttt{biodegradation}} & rec && 85.01  & 183 & 12 && 85.33 &  326 & 30 &&  85.64 & 571 & 33 & & & 100.0 & 76331 & 50& 47\\
                            & ret && 85.01  & 183 & 12 && 83.28 &  192 & 12 &&  84.06 & 194 & 12 & &  &  90.85 & 339 & 28&-\\
    \cmidrule{1-2} \cmidrule{4-6} \cmidrule{8-10} \cmidrule{12-14} \cmidrule{17-20}

        \multirow{2}{*}{\texttt{australian}} & rec && 84.54  & 118 & 10 && 85.02&  225 & 18 &&  85.50 & 302 & 19 & & $\cdots$ & 100.0 & 2527 & 24 & 30\\ 
                       & ret && 84.54  & 118 & 10 && 89.13 &  94 & 10 && 89.85 & 94 & 9 & & &  97.10 & 221 & 15&-\\
    \cmidrule{1-2}  \cmidrule{4-6} \cmidrule{8-10} \cmidrule{12-14} \cmidrule{17-20}

        \multirow{2}{*}{\texttt{bupa}} & rec && 92.30  & 52 & 9 && 93.26 &  81 & 12 &&  94.23 & 90 & 13 & & & 100.0 & 242 & 16& 8\\
                  & ret && 92.30  & 52 & 9 && 89.42 &  48 & 7 && 87.98 & 49 & 7 & & & 92.30 & 78 & 10&-\\
    \cmidrule{1-2}  \cmidrule{4-6} \cmidrule{8-10} \cmidrule{12-14} \cmidrule{17-20}

          \multirow{2}{*}{\texttt{german}} & rec && 93.33  & 76 & 7 && 93.66 &  117 & 15 &&  94.0 & 157 & 16 & & & 100.0 & 1099 & 21& 20\\ 
               & ret && 93.33 & 76 & 7 &&  95.33 &  77 & 8 &&  94.83 & 75 & 8 & & &  97.66 & 105 & 9&-\\
    \cmidrule{1-2}  \cmidrule{4-6} \cmidrule{8-10} \cmidrule{12-14} \cmidrule{17-20}

        \multirow{2}{*}{\texttt{contraceptive}} & rec && 71.49  & 676 & 19 && 71.94 &  663 & 19 &&  72.39 & 681 & 19 & & $\cdots$ & 100.0 & 4543 & 25 & 113\\
                           & ret && 71.49  & 676 & 19 && 85.29 &  392 & 14 && 86.19 & 385 & 14 & & & 94.57 & 559 & 23&-\\
    \cmidrule{1-2}  \cmidrule{4-6} \cmidrule{8-10} \cmidrule{12-14} \cmidrule{17-20}
         \multirow{2}{*}{\texttt{cleveland}} & rec && 85.71 & 72 & 7 && 87.36 &  101 & 10 &&  89.01 & 113 & 10 & & & 100.0 & 336 & 13 & 12\\
                       & ret && 85.71  & 72 & 7 && 89.01 &  61 & 7 &&  86.81 & 65 & 7 & & & 95.60 & 91 & 8&-\\  

         \cmidrule{1-2}  \cmidrule{4-6} \cmidrule{8-10} \cmidrule{12-14} \cmidrule{17-20}
                  
    \multirow{2}{*}{\texttt{compas}} & rec && 86.28  & 1022 & 16 && 87.87 &  550 & 15 &&  87.93 & 554 & 15 & & & 100.0 & 465 & 15& 71\\
                   & ret && 86.28  & 1022 & 16 && 99.62 &  314 & 12 && 99.59 & 314 & 12 & & & 99.64 & 341 & 12&-\\
    \cmidrule{1-2}  \cmidrule{4-6} \cmidrule{8-10} \cmidrule{12-14} \cmidrule{17-20}               
    \multirow{2}{*}{\texttt{cnae}} & rec && 98.61  & 56 & 13 && 98.91 &  34 & 8 &&  99.22 & 42 & 8 & & & 100.0 & 79 & 9& 5\\
                   & ret && 98.61  & 56 & 13 && 99.69 &  45 & 13 && 99.38 & 51 & 13 & & & 99.69 & 53 & 13&-\\
    \cmidrule{1-2}  \cmidrule{4-6} \cmidrule{8-10} \cmidrule{12-14} \cmidrule{17-20}               
    \multirow{2}{*}{\texttt{breast-tumor}} & rec && 64.53  & 96 & 7 && 65.69 &  99 & 11 &&  66.86 & 111 & 12 & & & 100.0 & 799 & 17& 30\\
                   & ret && 64.53  & 96 & 7 && 73.25 &  88 & 7 && 72.67 & 86 & 7 & & & 77.90 & 103 & 7&-\\
\cmidrule{1-2}  \cmidrule{4-6} \cmidrule{8-10} \cmidrule{12-14} \cmidrule{17-20} 
\multirow{2}{*}{\texttt{balance\_0\_vs\_1}} & rec && 90.69  & 164 & 12 && 91.22 &  168 & 12 &&  92.02 & 174 & 12 & & & 100.0 & 201 & 14 & 18\\
                          & ret && 90.69   & 164 & 12 && 92.28 &  154 & 13 &&  92.55 & 154 & 12 & & & 94.14 & 163 & 12 &-\\
\cmidrule{1-2}  \cmidrule{4-6} \cmidrule{8-10} \cmidrule{12-14} \cmidrule{17-20} 
\multirow{2}{*}{\texttt{balance\_0\_vs\_2}} & rec && 91.66  & 97 & 9 && 92.64 &  103 & 10 &&  93.62 & 107 & 10 & & & 100.0 & 135 & 9 & 8\\
                          & ret &&  91.66    & 97 & 9 && 92.64 &  96 & 9 &&  93.13 & 97 & 9 & & & 97.05 & 113 & 11 &-\\
\cmidrule{1-2}  \cmidrule{4-6} \cmidrule{8-10} \cmidrule{12-14} \cmidrule{17-20} 
\multirow{2}{*}{\texttt{balance\_1\_vs\_2}} & rec && 81.79  & 15 & 4 && 82.36 &  17 & 4 &&  84.68 & 23 & 5 & & & 100.0 & 111 & 9 & 25\\
                          & ret &&  81.79    & 15 & 4 && 82.08 &  15 & 4 &&  81.50 & 15 & 4 & & & 80.92 & 15 & 5 &-\\

    
                
\bottomrule
\end{tabular}}   
\caption{Empirical accuracy $I_P$ relative to $P$ (in \%), median number of nodes $N$, and median depth $D$ of decision trees for different datasets and each correction method when the decision trees are learned under the default configuration. $f$ is the median number of steps at which $T_I^{\pm}$ becomes empty when correction is performed by rectification.}
\label{tab:resultats-default}
\end{table}

\renewcommand{\arraystretch}{1.2}
\setlength{\tabcolsep}{2.5pt}
\begin{table}[H]
\centering        
\resizebox{\linewidth}{!}{\Large
\begin{tabular}{l rr rr r rr r rr r rr r rr r r rrr }
    \toprule
    \multicolumn{3}{c}{} & \multicolumn{16}{c}{Correction steps (optimized configuration)} \\
    \cmidrule{4-20}
    \multicolumn{2}{c}{Dataset} &&\multicolumn{3}{c}{0} &&\multicolumn{3}{c}{1}&&\multicolumn{3}{c}{2}&&&\multicolumn{4}{c}{Final step (rec)}  \\
    \cmidrule{4-6} \cmidrule{8-10} \cmidrule{12-14} \cmidrule{17-20}
    \multicolumn{3}{c}{} & $I_P$&$N$&$D$ &&   $I_P$&$N$&$D$   && $I_P$&$N$&$D$  && &$I_P$&$N$&$D$&$f$  \\ 
    \cmidrule{1-2}  \cmidrule{4-6} \cmidrule{8-10} \cmidrule{12-14} \cmidrule{17-20}
    \multirow{2}{*}{\texttt{bank}} & rec && 90.75  & 408 & 14 && 90.82 &  454 & 22 &&  90.89 & 426 & 25 & & & 100.0 & 9322 & 36 & 125\\
                  & ret && 90.75& 408 & 14 && 90.82 &  401 & 14 &&  91.04 & 403 & 14 & & &   94.03 & 560 & 14&-\\
    \cmidrule{1-2} \cmidrule{4-6} \cmidrule{8-10} \cmidrule{12-14} \cmidrule{17-20}

      \multirow{2}{*}{\texttt{biodegradation}} & rec && 86.75  & 30 & 3 && 87.06 &  100 & 29 &&  87.38 & 209 & 30 & & & 100.0 & 57826 & 47& 42\\
                            & ret && 86.75  & 30 & 3 && 85.80&  31 & 3 &&  83.28 & 31 & 3 & &  &   76.34 & 29 & 3&-\\
    \cmidrule{1-2} \cmidrule{4-6} \cmidrule{8-10} \cmidrule{12-14} \cmidrule{17-20}

        \multirow{2}{*}{\texttt{australian}} & rec && 87.43  & 45 & 4 && 87.92 &  69 & 14 &&  88.40 & 130 & 17 & & $\cdots$& 100.0 & 1477 & 22 & 25\\ 
                       & ret && 87.43  & 45 & 4 && 90.82 &  43 & 4 &&   91.06 & 36 & 4 & & &  79.46 & 43 & 4&-\\
    \cmidrule{1-2}  \cmidrule{4-6} \cmidrule{8-10} \cmidrule{12-14} \cmidrule{17-20}

        \multirow{2}{*}{\texttt{bupa}} & rec && 92.30  & 46 & 7 && 93.26 &  73 & 13 &&  94.23 & 79 & 13 & & & 100.0 & 266 & 14& 8\\

                  & ret && 92.30  & 46 & 7 && 87.01 &  46 & 7 && 86.53 & 44 & 7 & & &  86.53 & 59 & 7 &-\\
    \cmidrule{1-2}  \cmidrule{4-6} \cmidrule{8-10} \cmidrule{12-14} \cmidrule{17-20}
    
          \multirow{2}{*}{\texttt{german}} & rec && 95.5  & 44 & 4 && 95.83 &  57 & 13 &&  96.16 & 75 & 14 & & & 100.0 & 585 & 20& 13\\ 
               & ret && 95.5  & 44 & 4 && 97.5 &  41 & 4 &&  98.0 & 43 & 4 & & &  98.0 & 45 & 4&-\\
    \cmidrule{1-2}  \cmidrule{4-6} \cmidrule{8-10} \cmidrule{12-14} \cmidrule{17-20}

        \multirow{2}{*}{\texttt{contraceptive}} & rec && 82.57  & 24 & 3 && 82.80 &  39 & 11 &&  83.48 & 64 & 13 & &$\cdots$ & 100.0 & 1200 & 20 & 69\\
                           & ret && 82.57  & 24 & 3 && 87.33 &  22 & 3 && 86.42 & 23 & 3 & & & 86.19 & 29 & 3&-\\
    \cmidrule{1-2}  \cmidrule{4-6} \cmidrule{8-10} \cmidrule{12-14} \cmidrule{17-20}

         \multirow{2}{*}{\texttt{cleveland}} & rec && 86.81  & 29 & 3 && 88.46 &  36 & 6 &&  90.10 & 48 & 7 & & & 100.0 & 247 & 11 & 11\\
                       & ret && 86.81  & 29 & 3 && 89.56 &  29 & 3 &&  86.81 & 29 & 3 & & & 93.40  & 29 & 3&-\\    
        \cmidrule{1-2}  \cmidrule{4-6} \cmidrule{8-10} \cmidrule{12-14} \cmidrule{17-20}

    \multirow{2}{*}{\texttt{compas}} & rec && 93.25  & 63 & 4 && 93.60 &  45 & 6 &&  94.24 & 51 & 7 & & & 100.0 & 193 & 12& 38\\
                   & ret && 93.25  & 63 & 4 && 93.43 &  60 & 4 && 93.43 & 60 & 4 & & & 91.95 & 57 & 4&-\\
    \cmidrule{1-2}  \cmidrule{4-6} \cmidrule{8-10} \cmidrule{12-14} \cmidrule{17-20} 
    \multirow{2}{*}{\texttt{cnae}} & rec && 98.76  & 41 & 7 && 99.07 &  30 & 7 &&  99.38 & 37 & 7 & & & 100.0 & 65 & 8 & 5\\
                          & ret && 98.76  & 41 & 7 && 99.38 &  33 & 7 &&  99.69 & 35 & 7 & & & 99.69 & 41 & 7 &-\\
\cmidrule{1-2}  \cmidrule{4-6} \cmidrule{8-10} \cmidrule{12-14} \cmidrule{17-20} 
\multirow{2}{*}{\texttt{breast-tumor}} & rec && 66.27  & 148 & 14 && 67.44 &  159 & 14 &&  68.60 & 171 & 15 & & & 100.0 & 603 & 16 & 29\\
                          & ret && 66.27  & 148 & 14 && 74.41 &  141 & 14 &&  73.83 & 141 & 14 & & & 84.88 & 153 & 14 &-\\
\cmidrule{1-2}  \cmidrule{4-6} \cmidrule{8-10} \cmidrule{12-14} \cmidrule{17-20} 
\multirow{2}{*}{\texttt{balance\_0\_vs\_1}} & rec && 96.80  & 39 & 4 && 97.34 &  29 & 7 &&  97.87 & 41 & 7 & & & 100.0 & 65 & 9 & 6\\
                          & ret && 96.80  & 39 & 4 && 97.34 &  41 & 4 &&  97.60 & 40 & 4 & & & 96.27 & 39 & 4 &-\\
\cmidrule{1-2}  \cmidrule{4-6} \cmidrule{8-10} \cmidrule{12-14} \cmidrule{17-20} 
\multirow{2}{*}{\texttt{balance\_0\_vs\_2}} & rec && 93.62  & 24 & 3 && 94.60 &  31 & 7 &&  95.58 & 39 & 7 & & & 100.0 & 74 & 9 & 7\\
                          & ret && 93.62 & 24 & 3 && 94.11 &  25 & 3 &&  94.11 & 25 & 3 & & & 90.68 & 25 & 3 &-\\
\cmidrule{1-2}  \cmidrule{4-6} \cmidrule{8-10} \cmidrule{12-14} \cmidrule{17-20} 
\multirow{2}{*}{\texttt{balance\_1\_vs\_2}} & rec && 91.32  & 115 & 8 && 91.90 &  121 & 8 &&  92.48 & 127 & 8 & & & 100.0 & 197 & 8 & 16\\
                          & ret && 91.32 & 115 & 8 && 92.48 &  118 & 8 &&  92.77& 118 & 8 & & & 96.53 & 138 & 8 &-\\
                            \bottomrule
\end{tabular}}
\caption{Empirical accuracy $I_P$ relative to $P$ (in \%), median number of nodes $N$, and median depth $D$ of decision trees for different datasets and each correction method when the depth of the decision trees has been optimized. $f$ is the median number of steps at which $T_I^{\pm}$ becomes empty when correction is performed by rectification.}
\label{tab:resultats}
\end{table}
Table \ref{tab:resultats-default}  provides, for each dataset, the median values of 
$I_P$ (in \%) 
and the median number of nodes $N$ of the corrected tree for each of the two correction methods (rec is rectification, ret is retraining).
%
%
Table \ref{tab:resultats} provides the same type of information as Table \ref{tab:resultats-default}  when the depths of the decision trees have been optimized. 

In Tables \ref{tab:resultats-default} and \ref{tab:resultats}, a few correction steps are highlighted. Step $0$ is about the initial decision tree, step $1$ (resp. $2$, $f$) is about the decision tree obtained after $1$ (resp. $2$, $f$) correction steps.

Whatever the configuration used for learning the trees, after each rectification, \emph{it is guaranteed that $I_P$ strictly increases} since at least the instance triggering the correction step has been corrected at each step. The number $f$ of steps required to correct every instance of $T_I^{\pm}$ by rectifying it is thus well-defined. Comparing this number $f$ to the number $|T_I^{\pm}|$ of instances that were initially misclassified  (see Table \ref{description}), one may observe in Tables \ref{tab:resultats-default} and \ref{tab:resultats} that rectification with rules covering a possibly large number of instances can have a very significant impact on the number of correction steps required for achieving a complete distillation over $T$ (for example, for {\tt compas} under optimized configuration, 39 steps have been sufficient while 128 instances from $T$ were initially misclassified).

Contrastingly, the empirical results show that successive corrections by retraining can lead to decrease the 
value of $I_P$, 
making a \emph{complete correction impossible} in general. 
This holds under the optimized configuration, but also under the default configuration. In this last case, one knows that adding $(\vec x, P(\vec x))$ to the training set before retraining $I$ is enough to guarantee 
that $\vec x$ will be classified as required by $P$ after retraining. However, this is not enough to ensure that the accuracy of the decision tree relative to the boosted tree increases gradually with the correction steps. Indeed, a misclassified instance corrected at a certain step by retraining \emph{may become misclassified again after subsequent retraining steps}. 

\medskip
Tables \ref{tab:resultats-default} and \ref{tab:resultats} also show that the growth of the size of the trees (and of their depth) is more significant when rectification is used than when retraining is used, and that the size of rectified trees $I$ can become quite large, thus possibly questioning the benefits offered by the distillation process.
The empirical results presented in Table \ref{tab:growth} show that in spite of the growth of the size of the trees,\footnote{Notably, for each dataset, the $10$ decision trees one started with correspond to the default configuration. As reflected by the values of columns $f$, $N$, and $D$ in Tables \ref{tab:resultats-default} and \ref{tab:resultats}, this choice was less favourable than the choice of the optimized configuration in terms of numbers of rectification step and of size and depth of the tree $I$ obtained once all the rectifications have been achieved.} the distillation of $P$ into a decision tree $I$ is useful when the XAI objective that is pursued is to compute sufficient reasons.
In Table \ref{tab:growth}, $t_C$ is the average distillation time (in seconds) over the $10$ trees. For each tree, the distillation time includes the time required to identify instances $\vec x$ that are classified differently by the current decision tree $I$ and by $P$, the time required to compute an abductive explanation $t$ for each such $\vec x$ given $P$, and the time needed to rectify the current $I$ by the classification rule corresponding to $t$ (see Proposition \ref{prop:decision}).\footnote{Accordingly, it can be observed that $t_C$ is larger than the average cumulated rectification time $d_{\mathit{rec}}$ pointed out in Table \ref{description_time}.}
In Table \ref{tab:growth}, $t_I$ and $t_P$ are respectively the mean computation times
(in seconds) achieved by the algorithms for computing a sufficient reason over the set of instances (only instances for which the computation of a sufficient reason terminated in due time have been considered when evaluating the average). $to_I$ is the mean number of timeouts for the $10$ trees and $to_P$ the number of timeouts reached by the algorithms for computing a sufficient reason over the 100 instances. 
In Table \ref{tab:growth}, the improvement ratio $\alpha = \frac{t_{P}}{t_{I}}$ measures how many times faster is, on average, the computation of a sufficient reason when based on $I$ instead of $P$, while $\beta = \lceil \frac{t_C}{t_{P} - t_{I}}\rceil$, referred to as the break-even point, indicates (when positive) the number of explanation queries from which the compilation effort is compensated. 

As Table \ref{tab:growth} points it out, significant computational benefits about the derivation of sufficient reasons can be achieved by distilling $P$ into $I$. On the one hand, no timeouts have been observed when the computation of sufficient reasons was based on $I$, whatever the dataset and the decision tree at start: for each instance, a sufficient reason has been computed in less than 180 seconds. In contrast, the number of timeouts reached when sufficient reasons were derived from $P$ varies with the dataset and can be very significant (exceeding 50\% of the instances tested for the datasets {\tt bank} and {\tt contraceptive}). On the other hand, the improvement ratio $\alpha$ was huge (more than two orders of magnitude for every dataset). The break-even point $\beta = 1$ was equal to its least possible value  when $\alpha > 1$ whatever the dataset. This means that the distillation of $P$ into $I$ is valuable even when the computation of a sufficient reason is required for a single instance.
For more details,
we refer the reader to the supplementary material available from \cite{swh-dir-0689cda}.
 %
Of course, it must be kept in mind that the decision tree $I$ obtained after $f$ rectification steps is not equivalent to $P$ in general (it is only equivalent to $P$ on $T$), which limits the scope of the comparison. However, this is consistent with our objective (see Section \ref{sec:distilling}), since our goal is not to fully distill $P$ into a decision tree: in our approach, the distillation process is achieved incrementally, in a lazy and opportunistic fashion.  

\begin{table}[h!]
\begin{center}
\begin{tabular}{lccccccc}
\hline
Dataset & $t_C$ & $to_{I}$ & $t_{I}$ & $to_{P}$ & $t_{P}$ & $\alpha$ & $\beta$\\
\hline
{\tt bank}         & 39.74 & 0 & 0.62  & 54 & 102.25 & 164.91   & 1   
\\
{\tt biodegradation}   & 22.84 & 0 & 0.55  & 20 & 87.95  & 156.27   & 1   
\\
{\tt australian}       & 3.12  & 0 & 0.02  & 0  & 20.21  & 1010.5   & 1   
\\
{\tt bupa}             & 0.25  & 0 & 0.001 & 0  & 1.50   & 1500     & 1    \\
{\tt german}           & 0.61  & 0 & 0.012 & 0  & 8.02   & 668.33   & 1   \\
{\tt contraceptive}    & 4.84  & 0 & 0.02  & 85 & 71.68  & 3634   & 1    \\
{\tt cleveland}        & 0.56  & 0 & 0.005 & 0  & 5.24   & 1048     & 1    \\
{\tt compas}           & 4.95  & 0 & 0.007 & 0  & 23.31  & 3330  & 1    \\
{\tt cnae}             & 0.019 & 0 & 0.0003& 0  & 0.05   & 166.66   & 1    \\
{\tt breast-tumor}     & 0.79  & 0 & 0.004 & 19 & 78.26  & 19565    & 1  \\
{\tt balance\_0\_vs\_1} & 0.19  & 0 & 0.001 & 0  & 0.95   & 950      & 1     \\
{\tt balance\_0\_vs\_2} & 0.07  & 0 & 0.0009& 0  & 0.20   & 222.22   & 1    \\
{\tt balance\_1\_vs\_2} & 0.13  & 0 & 0.001 & 0  & 0.30   & 300      & 1    \\
\bottomrule

\end{tabular}
\end{center}
\caption{Evaluation of the computational benefits of distilling $P$ into $I$ in the objective of deriving sufficient reasons: compilation time $t_C$, numbers of timeout $to_{P}$ and $to_{I}$, mean
computation times $t_{P}$ and $t_{I}$, 
improvement ratio $\alpha$, and break-even point $\beta$.}
\label{tab:growth}
\end{table}

\begin{table}[h]
\begin{center}
\begin{tabular}{ l r r r r}
\toprule
Dataset & $d_{rec}$ & $d_{ret}$ & $o_{rec}$ & $o_{ret}$ \\
\midrule
{\tt bank} & 13.22 & 7978.16 & 4.18 & 3141.64\\
{\tt biodegradation} & 8.01 & 367.46 & 3.59 &196.69 \\
{\tt australian} &  0.65& 80.67& 0.047 & 44.69\\
{\tt bupa} & 0.007 & 4.29 & 0.003 & 2.78\\
{\tt german} & 0.048 & 29.59 & 0.008 & 8.87\\
{\tt contraceptive}  & 1.37 & 860.04 & 0.09 & 132.35\\
{\tt cleveland}  & 0.007 & 9.44 & 0.002 &  6.73\\
{\tt compas} & 0.09 & 142.01 & 0.012 & 35.65  \\
{\tt cnae} & 1.2e-3 & 0.54 & 5.5e-4 &  0.19  \\
{\tt breast-tumor} & 0.03 & 2.74 & 0.03 &  2.46  \\
{\tt balance\_0\_vs\_1} & 8.8e-3 & 1.06 & 1.6e-3 & 0.34   \\
{\tt balance\_0\_vs\_2} & 4.2e-3 & 0.29 & 4.7e-3 &  0.59  \\
{\tt balance\_1\_vs\_2} & 5.5e-3 & 2.46 & 8.3e-3 &  1.04  \\

\bottomrule
\end{tabular}
\end{center}

\caption{Comparison of the cumulative average computation times (in seconds) required by the two correction methods: (rec)tification and (ret)raining, in the default case (d) and in the optimized case (o).}
\label{description_time}
\end{table}

\medskip
Table \ref{description_time} gives the cumulative average times (in seconds) required by the successive correction operations up to the final rectification step $f$. $d_{rec}$ and $d_{ret}$ represent respectively the cumulative times for rectification and retraining in the default case, while $o_{rec}$ and $o_{ret}$ represent respectively the cumulative times for rectification and retraining in the optimized case. The results show that the computation times required to achieve the rectification process is small enough (and often two orders of magnitude smaller than the corresponding times when retraining is used). This shows the proposed rectification-based approach to distillation practical enough, despite the growth of the decision tree.
\begin{figure}[h!]
    \centering
    \includegraphics[width=0.75\linewidth]{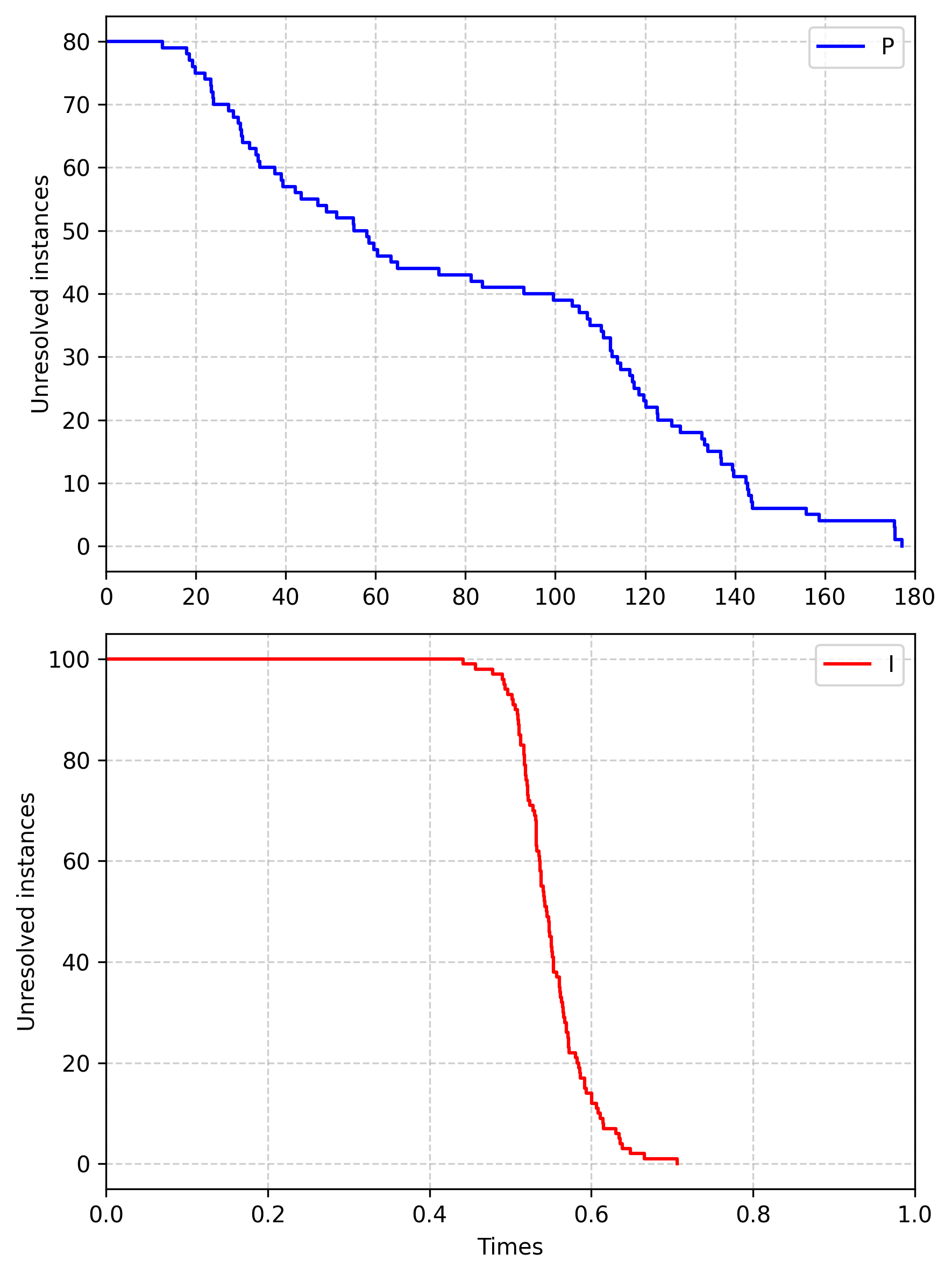}
    \caption{Comparison of unresolved instances over time for  the {\tt biodegradation} dataset.}
    \label{fig:unresolved_instances_biodegradation}
\end{figure}

Focusing on the {\tt biodegradation} dataset, Figure~\ref{fig:unresolved_instances_biodegradation} (top) reports the number of unresolved instances for the problem of generating explanations, i.e., the number of instances $\vec x$ for which the computation of a sufficient reason for $\vec x$ given $P$ within a given time limit fails. Only the $80$ instances for which the computation terminated within $180$ seconds were  considered at start. More precisely, the curve represents, at each computation time (in seconds) on the $x$-axis and viewed as a timeout value, the number of instances on the $y$-axis for which no sufficient reason has been derived 
within the time given on the $x$-axis. Figure~\ref{fig:unresolved_instances_biodegradation} (bottom) presents similar results when $I$ is used instead of $P$. One starts this time with the full set of $100$ instances since the computation of a sufficient reason terminated within $180$ seconds for each of them. We do not report on the curves the number of unresolved instances as soon as it becomes equal to zero.
As expected, for $P$ and for $I$, while time progresses, the number of unresolved instances  decreases. In less than $1$ second, a sufficient reason has been computed from $I$ for every instance from the set of $100$ instances one started with. In contrast, the computation of a sufficient reason from $P$ failed for every instance when the time allocated to the computation of a reason was ten times larger (i.e., 10 seconds).
Accordingly, the curve corresponding to $I$ is below the one corresponding to $P$, indicating that the computation of a sufficient reason from $I$ is more efficient than the computation of a sufficient reason  from $P$. A similar performance shift can be observed for every dataset considered in the experiments. Figure~\ref{fig:unresolved_instances_biodegradation} thus illustrates the clear advantage of distilling $P$ into $I$ when the goal is to guarantee a rapid  generation of explanations. Similar figures corresponding to the other datasets considered in the paper can be found in \cite{swh-dir-0689cda}.

\begin{figure}[t]
    \centering
    \centering\includegraphics[width=1\linewidth]{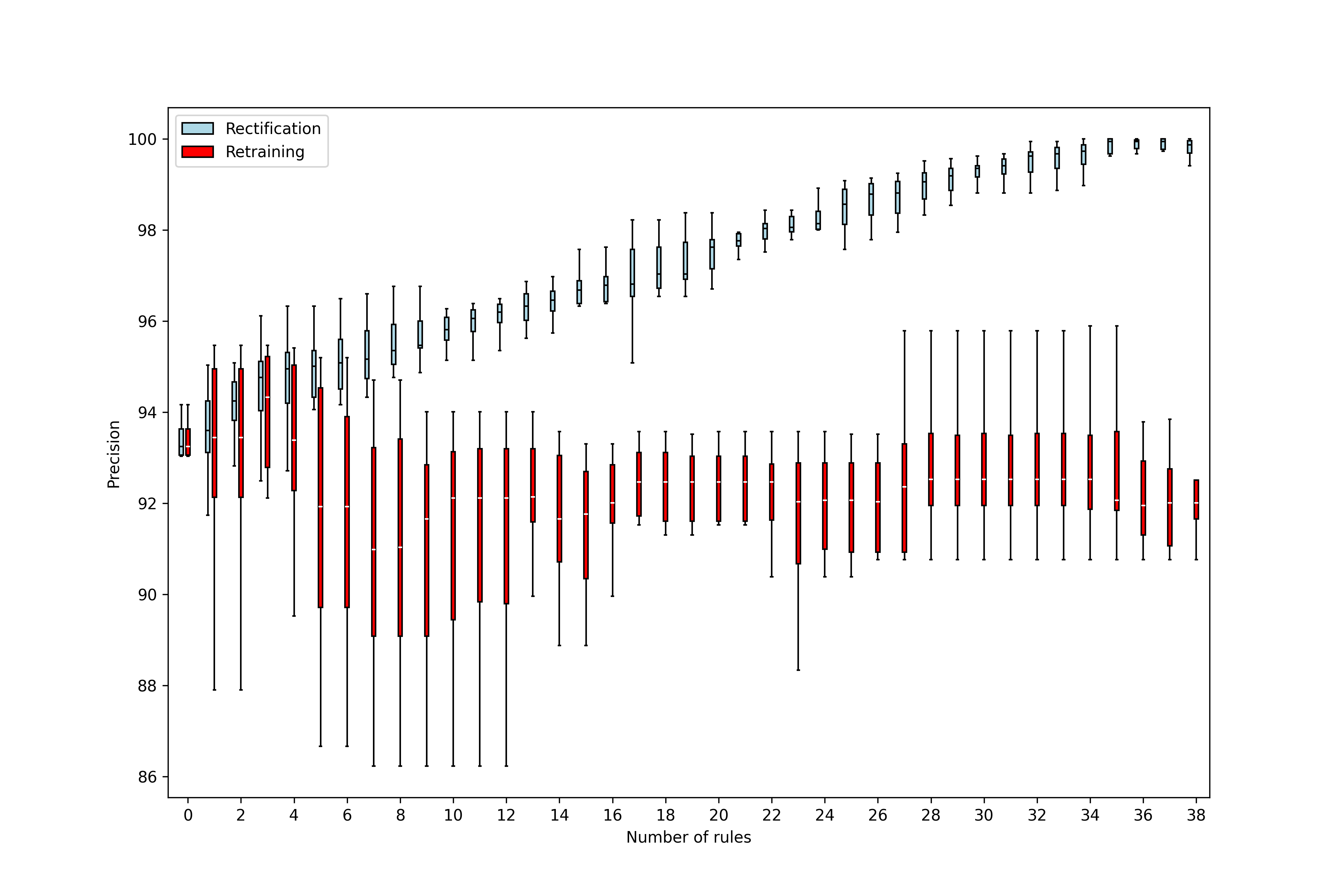}
    \caption{Relative accuracy $I_P$ obtained after rectification or retraining for 
 the {\tt compas} dataset under the optimized configuration.}
    \label{fig:compas_precision}
\end{figure}

\begin{figure}[t]
    \centering\includegraphics[width=1\linewidth]{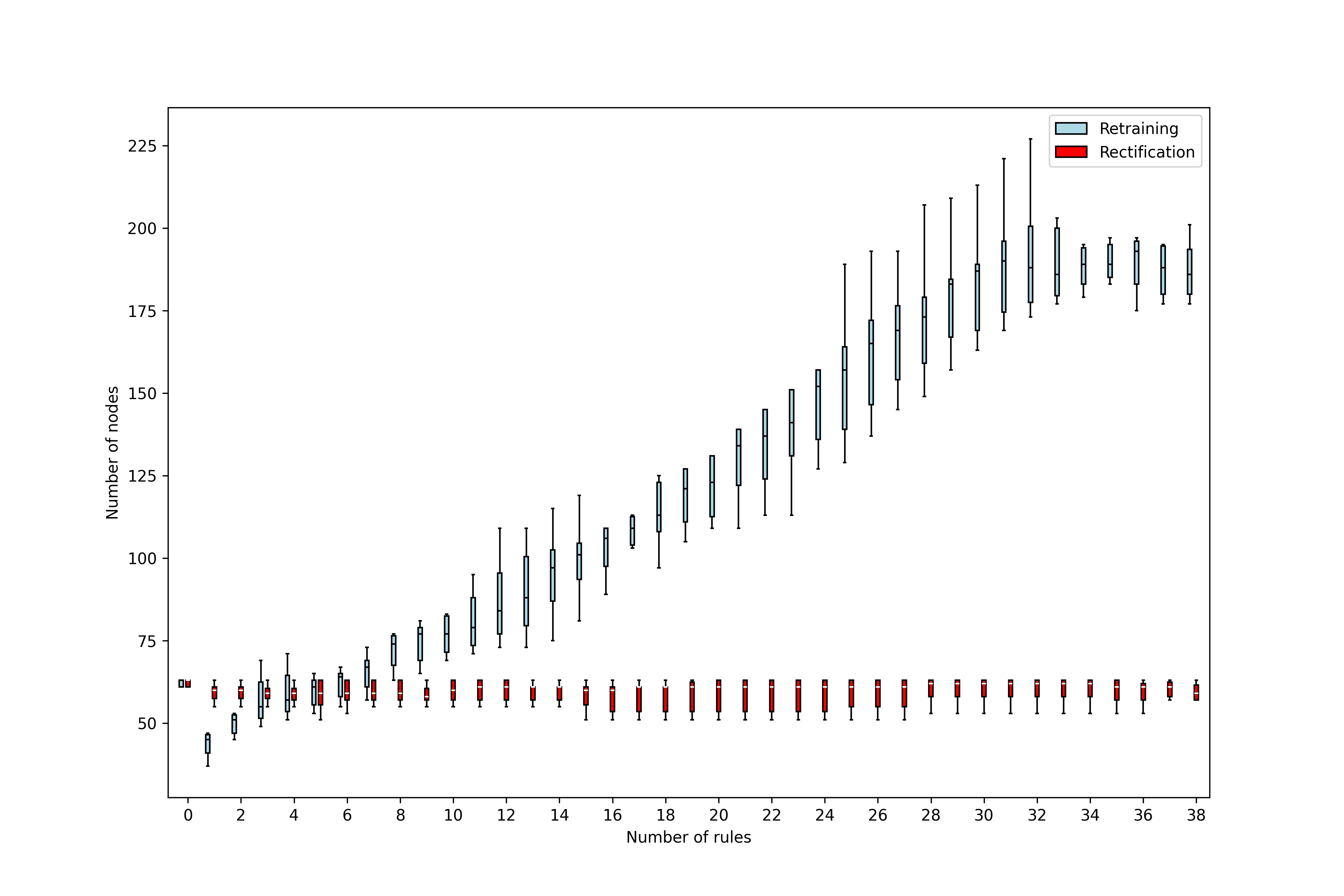}
    \caption{Number of nodes in the decision tree obtained after rectification or retraining for the {\tt compas} dataset under the optimized configuration.}
    \label{fig:compas_nb_noeuds}
\end{figure}


\medskip

Figure \ref{fig:compas_precision} and Figure \ref{fig:compas_nb_noeuds} provide more detailed statistics regarding the empirical results obtained for 
 the {\tt compas} dataset under the optimized configuration. The distributions of accuracy and tree sizes are summarized in box plots. 
The results obtained on {\tt compas} cohere with those given in Table \ref{tab:resultats}: the figure highlights the increase in relative accuracy yielded  by rectification (compared to retraining) at the cost of an increase of the size of the decision tree, which remains manageable nevertheless (the median size of the trees after 39 rectifications is (roughly) at most seven times the initial median size).

The accuracy/size trade-off shown for {\tt compas} can be observed for the other data\-sets. Similar figures corresponding to the other datasets considered in the paper can be found in \cite{swh-dir-0689cda}.

\section{Other Related Work}

In ML, a number of approaches have been designed so far to tackle the problem of explaining tree ensemble models to mitigate the trust risks associated with their lack of transparency, see e.g.,  \cite{FriedmanPopescu08,Deng18,DBLP:journals/inffus/ObregonJ23}. Most of the time, sets of classification rules are targeted (even if, in some approaches \cite{DBLP:journals/isci/SagiR21,DBLP:journals/inffus/SagiR20}, those sets of rules are finally combined into decision trees). However,  sets of rules (alias \cnf\ classifiers) are not computationally intelligible \cite{DBLP:conf/kr/AudemardBBKLM21}, unless some constraints are imposed on the number or on the types of rules. But the presence of such constraints then prevents the full distillation of the tree-based model from being possible (i.e., the resulting set of rules approximates the tree-based model). Similarly, in the general case, a set of classification rules cannot be turned into an equivalent decision tree for expressiveness reasons: the former corresponds in general to an incomplete classifier, while a decision tree always is a complete classifier. 
Alternatively, \cite{LI2025103244} shows how to turn tree ensembles into decision trees without considering sets of rules. The resulting trees approximate the tree ensembles considered as input. Experiments have shown that the approximation achieved can be of good quality, but, as mentioned by the authors, the scalability of the proposed approach to high-dimensional datasets remains a challenge.


\medskip
Closer to our approach is the “Born-Again Tree Ensembles (BATE)” approach presented in \cite{DBLP:conf/icml/VidalS20}, and more specifically, the heuristic approach reported in this paper. This heuristic approach aims to compute a decision tree that is \emph{equivalent} to a given weighted random forest, thus offering the very same guarantees as our own approach.
Beyond the fact that the inputs considered in the two approaches are not identical (weighted random forests vs. boosted trees), the methods at work to generate decision trees (dynamic programming vs. successive rectifications) in the two approaches are completely different. Furthermore, no data preprocessing is required in our approach (while, in BATE, continuous features are binned into 10 ordinal scales). Another significant difference is that our approach is lazy and opportunistic (which is not the case of BATE). This difference may explain the large discrepancy we observed empirically between the two approaches in terms of scalability. Indeed, we ran  heuristic BATE (see \url{https://github.com/vidalt/BA-Trees}) on our datasets for random forests with  a reduced number of trees (10 trees) of limited depth and parameter -obj set to 4, considering a timeout of 4 hours. heuristic BATE succeeded in computing in due time a decision tree equivalent to the input random forest only for two datasets ({\tt contraceptive} - the resulting tree contains 281 752 nodes and it has been computed in 174 seconds - and {\tt compas} - the resulting tree contains 7 806 968 nodes and it has been computed in 2248 seconds). A timeout occurred for all the remaining datasets. We also tried to increase the number of trees and/or the tree depth in order to increase the predictive performance of the forest one started with but under such conditions, heuristic BATE failed to return a decision tree within 4 hours. 

\medskip
In other approaches to the distillation of tree ensembles into decision trees (see e.g., \cite{BreimanShang96} \cite{Zhouetal23}),
the transformation of the input tree ensemble into a decision tree relies on an \emph{improved learning step}: the black box classifier $P$ is simply used as an oracle for generating new instances in order \emph{to complete the training set used to learn $I$}. No instances are removed from the training set. Thus, a decision tree $I$ is built up once for all (no correction of it is to be done once it has been derived). The issues to be considered for computing $I$ are the generation of new instances (connected to the splitting rule used), the stopping rule (deciding when the splitting process must be stopped), and the pruning rule. A common principle guiding the generation of new instances (``manufacturing data'' \cite{BreimanShang96}) is to ensure to have sufficiently many instances at each decision node to decide when to split and how to split. Those additional instances are useful to stabilize the greedy splitting strategy that is leveraged to construct the decision tree, leading to improve its accuracy \cite{Zhouetal23}. In order to satisfy those conditions, the number of extra instances to be generated is typically huge, leading to high learning times. Despite of it, there is in general no guarantee that the predictive performance of the decision tree that is learned is close to the one of $P$.

\medskip
Under the correction by re-training approach, an update of the training set used to learn $I$ is also considered but 
it is guided by the classification rule $R$ deduced from $P$ and based on the way $P$ classify $\vec x$. Indeed, all the instances $(\vec x', c)$ such that the rule $t_{\vec x'} \Rightarrow c$ and the rule $R$ are conflicting are removed from the training set, while some other instances $(\vec x', P(\vec x))$ such that $\vec x'$ is covered by the premises $t$ of $R$ are added.

\medskip
In contrast to all the approaches mentioned above to distilling a tree ensemble into a decision tree, our approach is \emph{correction-guided}. Rectification is leveraged to correct the current decision tree in an opportunistic way. No new instance is to be generated but only one explanation for each instance $\vec x$ that triggers a correction. Furthermore, a key feature of rectification its that \emph{it offers logical guarantees}: it ensures that the corrections that are targeted are effective, which is not the case of distillation approaches based on a completion of the training set and/or on (re)training. Especially, the trees resulting from such tree distillation approaches are not guaranteed to be equivalent to the black box considered at start, while this is ensured by our approach ``in the limit'', i.e., provided that all instances to be corrected are rectified at some step.


\section{Conclusion}

We have presented a new approach for distilling a boosted tree $P$ into a decision tree $I$. Our approach is based on the rectification operation that provides guarantees that the corrections that are requested are effective.
In contrast to a distillation approach that would boil down to 
translating $P$ into $I$ in one step, 
our approach is lazy and opportunistic: $I$ is corrected only when an instance $\vec x$ encountered at inference time is such that $I(\vec x) \neq P(\vec x)$. 
This makes it possible to control
the size of the resulting (rectified) decision tree $I$, therefore 
enabling to stop the rectification process as soon as desired.
Experiments have shown that our approach to distillation may provide interesting results compared to previous approaches.

\medskip
Interestingly, algorithms for minimizing decision trees could be leveraged within the proposed approach. Basically, the idea would be to interleave minimization steps with correction steps so as to increase the number of correction steps that could be handled while ensuring that the size of the resulting decision tree remains ``small enough''.
However, since minimizing a decision tree is {\sf NP}-hard \cite{DBLP:journals/jcss/Sieling08}, taking advantage of minimization steps could have a significant impact on the time needed to achieve the rectification process. An empirical evaluation will be necessary to determine the extent to which such an approach could be practical enough. As a alternative, distilling boosted trees into structured \dDNNF\ circuits \cite{PipatsrisawatDarwiche08} can be considered as well. Indeed, such circuits appear as interesting targets for a distillation process since they can be more succinct than decision trees, they support many XAI queries \cite{DBLP:conf/kr/AudemardBBKLM21}, and in light of Proposition \ref{prop:simpl}, they can also be rectified in polynomial time by classification rules. Evaluating the benefits that can be obtained by distilling a boosted tree into a structured \dDNNF\ circuit is another perspective for further research.





\bibliographystyle{splncs04}
\bibliography{kr-sample}

\newpage

\section*{Appendix: Proofs}

\smallskip

\noindent {\bf Proof of Proposition \ref{prop:simpl}}
\begin{proof}
It is easy to verify that each classification rule $R = \varphi_X \Rightarrow y$ (resp. $R = \varphi_X \Rightarrow \overline{y}$) satisfies $R(y) \equiv \top$ and $R (\overline{y}) \equiv \neg \varphi_X$ (resp. $R(\overline{y}) \equiv \top$ and $R(y ) \equiv \neg \varphi_X$). It is then sufficient to replace $R(y)$ and $R(\overline{y})$ by the equivalent expressions above in the definition of $\Sigma \star T$ and to simplify the result obtained to obtain the characterization given in the proposition.
\end{proof}


\noindent {\bf Proof of Proposition \ref{prop:all}}
\begin{proof}
The proof is based on three lemmas. The first lemma shows that the classification rules that can be deduced from a classification circuit on $X \cup \{y\}$ are never conflicting (two classification rules are conflicting whenever they have consistent premises and distinct conclusions).

\begin{lemma}\label{prop:notconflict}
Let $\Sigma = \Sigma_X \Leftrightarrow y$ be a classification circuit on $X \cup \{y\}$.
Two classification rules $R_1 = \varphi_X^1 \Rightarrow y$ and $R_2 = \varphi_X^2 \Rightarrow \overline{y}$ that can be deduced from $\Sigma$ are never conflicting.
\end{lemma}

\noindent {\it Proof.} $R_1 = \varphi_X^1 \Rightarrow y$ can be deduced from $\Sigma$ iff $\Sigma \wedge \varphi_X^1 \wedge \overline{y}$ is contradictory iff $(\Sigma_X \Leftrightarrow y) \wedge \varphi_X^1 \wedge \overline{y}$ is contradictory iff $\overline{\Sigma_X} \wedge \varphi_X^1$ is contradictory iff we have $\varphi_X^1 \models \Sigma_X$. Similarly, $R_2 = \varphi_X^2 \Rightarrow \overline{y}$ can be deduced from $\Sigma$ iff $\Sigma \wedge \varphi_X^2 \wedge y$ is contradictory iff $(\Sigma_X \Leftrightarrow y) \wedge \varphi_X^2 \wedge y$ is contradictory iff $\Sigma_X \wedge \varphi_X^2$ is contradictory iff we have $\varphi_X^2 \models \overline{\Sigma_X}$. Consequently, $\varphi_X^1 \wedge \varphi_X^2$ is necessarily contradictory.

\smallskip

The second lemma shows that if $R_1$ and $R_2$ are two classification rules over $y$ (resp. two classification rules over $\overline{y}$),
then rectifying a classifier $\Sigma$ by $R_1$ first and by $R_2$ then is equivalent to rectifying $\Sigma$ by the conjunction
$R_1 \wedge R_2$ of the two rules.

\begin{lemma}\label{prop:same}
Let $\Sigma = \Sigma_X \Leftrightarrow y$ be a classification circuit on $X \cup \{y\}$.
Let $R_1$ and $R_2$ be two classification rules over $y$ or two classification rules over $\overline{y}$.
We have $\Sigma \star (R_1 \wedge R_2) \equiv (\Sigma \star R_1) \star R_2$.
\end{lemma}

\noindent {\it Proof.} Let us first assume that $R_1 = \varphi_X^1 \Rightarrow y$ and $R_2 = \varphi_X^2 \Rightarrow y$ are two classification rules over $y$. We have $R_1 \wedge R_2 \equiv (\varphi_X^1 \vee \varphi_X^2) \Rightarrow y$, showing that $R_1 \wedge R_2$ is equivalent to the classification rule $\varphi_X \Rightarrow y$ over $ y$, with $\varphi_X \equiv (\varphi_X^1 \vee \varphi_X^2)$. Next, we exploit the fact that $\Sigma_X^{R_1 \wedge R_2}$ can be simplified to $\Sigma_X \vee \varphi_X^1 \vee \varphi_X^2$. Furthermore, we have $\Sigma_X^{R_1} \equiv \Sigma_X \vee \varphi_X^1$ and $\Sigma_X^{R_1, R_2} \equiv \Sigma_X^{R_1} \vee \varphi_X^2$. Since $\Sigma_X^{R_1} \equiv \Sigma_X \vee \varphi_X^1$, we have $\Sigma_X^{R_1 \wedge R_2} \equiv \Sigma_X^{R_1, R_2}$, which concludes the proof.

Similarly, suppose that $R_1 = \varphi_X^1 \Rightarrow \overline{y}$ and $R_2 = \varphi_X^2 \Rightarrow \overline{y}$ are two classification rules over $\overline{y}$. We have $R_1 \wedge R_2 \equiv (\varphi_X^1 \vee \varphi_X^2) \Rightarrow \overline{y}$, showing that $R_1 \wedge R_2$ is equivalent to the classification rule $\varphi_X \Rightarrow \overline{y}$ over $\overline{y}$, with $\varphi_X \equiv (\varphi_X^1 \vee \varphi_X^2)$. Next, we exploit the fact that $\Sigma_X^{R_1 \wedge R_2}$ can be simplified to $\Sigma_X \wedge \neg(\varphi_X^1 \vee \varphi_X^2)$, which is equivalent to $\Sigma_X \wedge \neg \varphi_X^1 \wedge \neg \varphi_X^2$. On the other hand, we have $\Sigma_X^{R_1} \equiv \Sigma_X \wedge \neg \varphi_X^1$ and $\Sigma_X^{R_1, R_2} \equiv \Sigma_X^{R_1} \wedge \neg \varphi_X^2$. Since $\Sigma_X^{R_1} \equiv \Sigma_X \wedge \neg \varphi_X^1$, we have $\Sigma_X^{R_1 \wedge R_2} \equiv \Sigma_X^{R_1, R_2}$, which concludes the proof.

\smallskip

Since conjunction is commutative, the resulting circuit is equivalent to
$\Sigma$ rectified by $R_2$ first and by $R_1$ then, that is, we have $\Sigma \star (R_1 \wedge R_2) \equiv (\Sigma \star R_2) \star R_1$.
In other words, the  rectification order does not matter.

\medskip

The third lemma concerns classification rules having contradictory premises and contradictory conclusions
($y$ and $\overline{y}$). For such rules $R_1$ and $R_2$, here again, rectifying a classification circuit $\Sigma$ by the conjunction $R_1 \wedge R_2$ amounts
to rectify $\Sigma$ by $R_1$ first and by $R_2$ then. And since conjunction is commutative, the rectification order actually does not matter.

\begin{lemma}\label{prop:different}
Let $\Sigma = \Sigma_X \Leftrightarrow y$ be a classification circuit on $X \cup \{y\}$.
Let $R_1 = \varphi_X^1 \Rightarrow y$ and $R_2 = \varphi_X^2 \Rightarrow \overline{y}$ be two classification rules such that $\varphi_X^1 \wedge \varphi_X^2 $ is contradictory.
We have $\Sigma \star (R_1 \wedge R_2) \equiv (\Sigma \star R_1) \star R_2$.
\end{lemma}

\noindent {\it Proof.}
On the one hand, we have 

$$\Sigma_X^{R_1 \wedge R_2} \equiv (\Sigma_X \wedge \neg ((R_1 \wedge R_2)(\overline{y}) \wedge \neg (R_1 \wedge R_2) (y))) \vee ((R_1 \wedge R_2)(y) \wedge \neg (R_1 \wedge R_2)(\overline{y})).$$


Since $(R_1 \wedge R_2)(\overline{y})$ is equivalent to $R_1(\overline{y}) \wedge R_2(\overline{y})$ and $(R_1 \wedge R_2)(y) $ is equivalent to $R_1(y) \wedge R_2(y)$, $\Sigma_X^{R_1 \wedge R_2}$ is equivalent to 

$$(\Sigma_X \wedge \neg (R_1(\overline{y}) \wedge R_2(\overline{y}) \wedge \neg (R_1(y) \wedge R_2(y)))).$$


Now, we exploit the facts that $R_1(y) \equiv R_2(\overline{y}) \equiv \top$, that $R_1(\overline{y}) \equiv \neg \varphi_X^1$, and that $R_2(y) \equiv \neg \varphi_X^2$ to simplify the previous formula. We obtain that $$\Sigma_X^{R_1 \wedge R_2} \equiv (\Sigma_X \wedge \neg (\neg \varphi_X^1 \wedge \varphi_X^2)) \vee (\varphi_X^1 \wedge \neg \varphi_X ^2).$$ Since $\varphi_X^1 \wedge \varphi_X^2$ is contradictory, we have $\neg \varphi_X^1 \wedge \varphi_X^2 \equiv \varphi_X^2$ and $\varphi_X^1 \wedge \neg \varphi_X ^2 \equiv \varphi_X^1$. Thus, $\Sigma_X^{R_1 \wedge R_2}$ is equivalent to $(\Sigma_X \wedge \neg \varphi_X^2) \vee \varphi_X^1$.

On the other hand, we have $\Sigma_X^{R_1} \equiv \Sigma_X \vee \varphi_X^1$ since $R_1$ is a classification rule over $y$. So, given that $R_2$ is a classification rule over $\overline{y}$, we have $\Sigma_X^{R_1, R_2} \equiv \Sigma_X^{R_1} \wedge \neg \varphi_X^2 \equiv ( \Sigma_X \vee \varphi_X^1) \wedge \neg \varphi_X^2 \equiv (\Sigma_X \wedge \neg \varphi_X^2) \vee (\varphi_X^1 \wedge \neg \varphi_X^2)$ . Since $\varphi_X^1 \wedge \neg \varphi_X^2 \equiv \varphi_X^1$, this last formula is equivalent to $(\Sigma_X \wedge \neg \varphi_X^2) \vee \varphi_X^1$. Therefore, $\Sigma_X^{R_1 \wedge R_2}$ is equivalent to $\Sigma_X^{R_1, R_2}$, and this concludes the proof.

\smallskip
Lemma \ref{prop:notconflict} shows that we can take advantage of Lemma \ref{prop:different}
whenever two rules $R_1 = \varphi_X^1 \Rightarrow y$ and $R_2 = \varphi_X^2 \Rightarrow \overline{y} $ are deduced from a classification circuit $\Phi_X \Leftrightarrow y$ where $\Phi_X$ is a binary classifier.
Indeed, Lemma \ref{prop:notconflict} ensures that for such rules, $\varphi_X^1 \wedge \varphi_X^2 $ is contradictory (otherwise, the two rules would conflict).

\smallskip
Finally, a simple induction on $k$ can be used to obtain the desired result from Lemma \ref{prop:same} and Lemma \ref{prop:different}.
\end{proof}


\noindent {\bf Proof of Proposition \ref{prop:decision}}
\begin{proof}
Let $R = t \Rightarrow y$. If $t$ is an abductive explanation for $\vec x$ given $C$, then we have $t \models C_X$. Equivalently, $\neg C_X \models \neg t$ holds. Since $y \not \in X$, this is equivalent to $\neg C_X \vee y \models \neg t \vee y$, i.e., $C_X \Rightarrow y \models t \Rightarrow y$. Since $\Phi \models C_X \Rightarrow y$, we have $\Phi \models R$. The case when $R = t \Rightarrow \overline{y}$ is similar.
\end{proof}

\noindent {\bf Proof of Proposition \ref{prop:converge}}

\begin{proof}
By construction, $(I_X \Leftrightarrow y) \star R$ is a classification circuit that classifies every instance $\vec x' \in \vec X$ as the classification circuit $I_X \Leftrightarrow y$, except those instances $\vec x'$ such that $I(\vec x') \neq R(\vec x')$, which are classified by $(I_X \Leftrightarrow y) \star R$ in the same way as they are classified by $R$ \cite{coste-marquis23}. Let us consider any instance $\vec x' \in \vec X_{I^R}^{\pm}$. Then $\vec x'$ is not covered by $R$, otherwise we would have $I^R(\vec x') = P(\vec x')$ given that $R$ is implied by the classification circuit $P_X \Leftrightarrow y$. Therefore, $\vec x'$ is classified by $I^R$ in the same way as it is classified by $I$, so that we also have $\vec x' \in \vec X_{I}^{\pm}$. To prove that the inclusion $X_{I^R}^{\pm} \subset X_I^{\pm}$ is strict, it is enough to observe that $R$ covers at least one instance that belongs to $X_I^{\pm}$, namely the instance $\vec x$ that triggered the correction step. This instance is classified by $I^R$ in the same way as it is classified by $R$, thus in the same way as it is classified by $P$ since $R$ is implied by the classification circuit $P_X \Leftrightarrow y$.
Accordingly, $\vec x \not \in X_{I^R}^{\pm}$, which completes the proof.
\end{proof}

\end{document}